\documentclass[lettersize,journal]{IEEEtran}
\usepackage{amsmath,amsfonts}
\usepackage{algorithmic}
\usepackage{algorithm}
\usepackage{array}
\usepackage[caption=false,font=normalsize,labelfont=sf,textfont=sf]{subfig}
\usepackage{textcomp}
\usepackage{stfloats}
\usepackage{url}
\usepackage{verbatim}
\usepackage{graphicx}
\usepackage{cite}

\usepackage{hyperref}       
\usepackage{booktabs}       
\usepackage{nicefrac}       
\usepackage[misc]{ifsym}
\usepackage{microtype}      
\usepackage[dvipsnames]{xcolor}
\usepackage{enumitem}
\setlist{nolistsep, leftmargin=0.1in}

\usepackage{amssymb}
\usepackage{multirow}
\usepackage{makecell}
\usepackage{caption}
\usepackage{tabularx}
\usepackage{xltabular}
\usepackage{longtable}

\usepackage{framed}
\newbox\totalbox
\newbox\partialbox
\newdimen\partialboxdim

\usepackage{chngcntr}
\begin{document}

\title{TinyLVLM-eHub: Towards Comprehensive and Efficient Evaluation for Large Vision-Language Models }

\author{Wenqi Shao$^{*1}$, {Meng Lei}$^{*1,4}${,} Yutao Hu$^{*1,2}$, Peng Gao$^{*1}$,  {Peng Xu}$^{1,2}${,} Kaipeng Zhang$^{1}$, \\ {Fanqing Meng}$^{1}${,}    {Siyuan Huang}$^{1}${,}    {Hongsheng Li}$^{1,3}${,}  {Yu Qiao}$^{1}$\textsuperscript{\Letter}{,} Senior Member, IEEE, {Ping Luo}$^{2, 1}$\textsuperscript{\Letter}
\thanks{$^{1}$ OpenGVLab, Shanghai AI Laboratory, Shanghai, China  $^2$Department of Computer Science, The University of Hong Kong, Hong Kong, China, $^{3}$Department of Electronic Engineering, The Chinese University of Hong Kong, Hong Kong, China $^{4}$Academy for Advanced Interdisciplinary Studies, Peking University, Beijing, China}
\thanks{$^*$ Equal First Authors}
\thanks{\textsuperscript{\Letter} Corresponding Authors: qiaoyu@pjlab.org.cn; pluo@cs.hku.hk }
\thanks{ Project Page: \url{http://lvlm-ehub.opengvlab.com/}}
}
\markboth{Journal of \LaTeX\ Class Files,~Vol.~14, No.~8, August~2021}%
{Shell \MakeLowercase{\textit{et al.}}: A Sample Article Using IEEEtran.cls for IEEE Journals}


\maketitle

\begin{abstract}

Large Vision-Language Models (LVLMs) have made significant strides in various multimodal tasks. Notably, GPT4V, Claude, Gemini, and others showcase exceptional multimodal capabilities, marked by profound comprehension and reasoning skills. This study introduces a comprehensive and efficient evaluation framework, TinyLVLM-eHub, to assess LVLMs' performance, including proprietary models. TinyLVLM-eHub covers six key multimodal capabilities, such as visual perception, knowledge acquisition, reasoning, commonsense understanding, object hallucination, and embodied intelligence. The benchmark, utilizing 2.1K image-text pairs, provides a user-friendly and accessible platform for LVLM evaluation. The evaluation employs the ChatGPT Ensemble Evaluation (CEE) method, which improves alignment with human evaluation compared to word-matching approaches. Results reveal that closed-source API models like GPT4V and GeminiPro-V excel in most capabilities compared to previous open-source LVLMs, though they show some vulnerability in object hallucination. This evaluation underscores areas for LVLM improvement in real-world applications and serves as a foundational assessment for future multimodal advancements. Find our project at \url{https://github.com/OpenGVLab/Multi-Modality-Arena}.
\end{abstract}


\section{Introduction}
\IEEEPARstart{L}{arge} Vision-Language Models (LVLMs) have demonstrated remarkable success in various multimodal applications, including visual complex reasoning \cite{zhu2023minigpt, li2023otter}, visual conversation \cite{llava, gao2023llama}, and medical visual question answering \cite{li2023llava, moor2023medflamingo}. The proliferation of various LVLMs has significantly advanced our comprehension and pushed the boundaries of multimodal applications \cite{zhang2023gpt4roi} across various domains \cite{moor2023medflamingo}. 

LVLM is typically constructed by incorporating a Large Language Model (LLM) with a pre-trained visual encoder which facilitates the integration of images as input data \cite{li2023blip, dai2023instructblip}. For example, LLaVA \cite{llava}, LLaMA-Adapter V2 \cite{gao2023llama}, and Otter \cite{li2023otter} feed LLM such as Vicuna \cite{vicuna} or LLaMA \cite{touvron2023llama} with visual tokens extracted by visual encoder ViT-L/14 \cite{fang2023eva}. Notably, closed-source models such as GPT4V \cite{gpt4} and GeminiPro-V \cite{geminiprov}, distinguish themselves with their exceptional multimodal capabilities. Although detailed model configurations are unknown, the open-source APIs of these proprietary models have demonstrated the multimodal ability. This groundbreaking development marks a significant step forward in the field of artificial general intelligence (AGI).

Despite the great success, it is vital to understand LVLMs' capabilities in various multimodal tasks. Recent work \cite{gvt, xu2023lvlm} attributes the success of LVLM to the representational power of the visual encoder, proper alignment between vision and language \cite{li2023blip, zhu2023minigpt}, and visual instructional tuning of LLM \cite{li2023mimic}. However, a comprehensive evaluation of LVLMs remains underdeveloped. Another line of research \cite{fu2023mme, liu2023mmbench, li2023seed} investigates various multimodal capabilities by experimenting with a large number of text-related visual benchmarks.  Nevertheless, these studies could not assess LVLMs' abilities in the open-set setting because they constrain the model's answer to be close-set options such as Yes/No and A/B/C/D. Moreover, they often comprise amounts of test samples while assessing a limited number of multimodal capabilities as shown in Table \ref{tab:comparison-LVLM-bench}, making it incomplete and cumbersome for evaluating LVLMs.


\begin{figure}[t]
		\centering
		\includegraphics[width=1.0\linewidth]{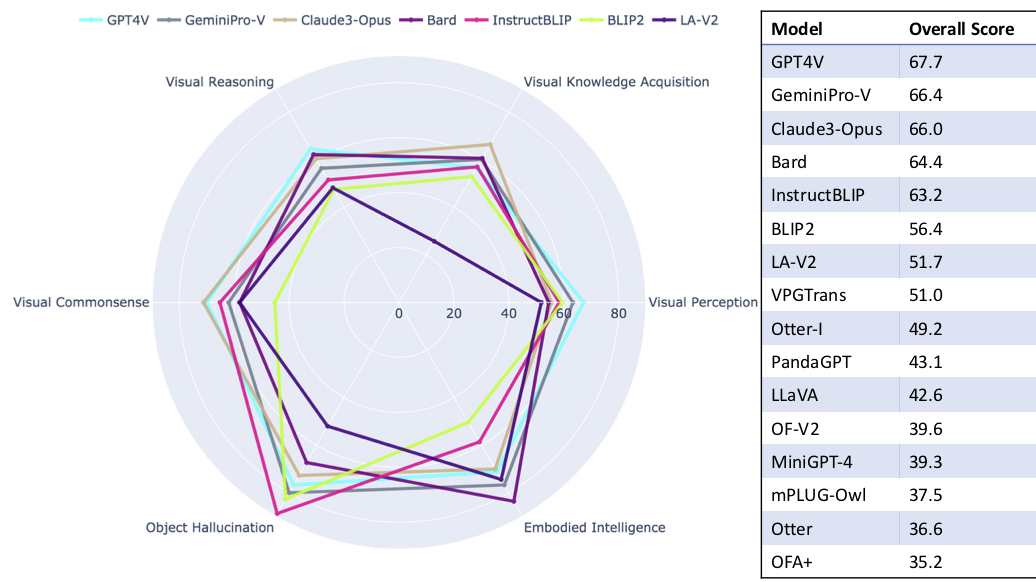}
		\caption{The overall multimodal score of $16$ LVLMs in TinyLVLM-eHub obtained by averaging over six capabilities. The capability scores of top-performing LVLMs are visualized. We see that closed-source API models such as GPT4V and GeminiPro-V perform well in our benchmark. However, they face more severe hallucination issues than the open-source InstrucBLIP model.
		}
		\label{fig:general1}
\end{figure}

In this work, we propose TinyLVLM-eHub to systematically evaluate various multimodal capabilities of numerous Large Vision-Language Models (LVLMs). Towards this goal, we consolidate 42 standard text-related visual benchmarks, from each of which $50$ examples are sampled. TinyLVLM-eHub is a lightweight version of LVLM-eHub \cite{xu2023lvlm} but enjoys several intrinsic properties. First, despite the simplicity, it can assess six categories of multimodal capabilities of LVLMs by inheriting advantages from LVLM-eHub as verified by Table \ref{tab:corr_lvlmehubs}. Second, TinyLVLM-eHub employs ChatGPT as a judge to assess the model's prediction and reference answer in an open-set scenario. By creating a prompt template with critical thinking, the resulting judgment aligns better with Human evaluation than the word-matching approach adopted by prior studies \cite{xu2023lvlm,liu2023hidden} (see Table \ref{tab:gpt_vs_hasword}). Third, LVLM-eHub has only 2.1K image-text pairs. Thus it is convenient for practitioners to evaluate their offline LVLMs on TinyLVLM-eHub. 

The comprehensive evaluation of TinyLVLM-eHub shows (see Fig. \ref{fig:general1}) that proprietary API-based models such as GPT4V and GemoniPro-V consistently outperform prior open-source LVLMs in various multimodal capabilities including visual perception, visual knowledge acquisition, visual reasoning, visual commonsense, and embodied intelligence, but still suffer from hallucination issues. The overall performance listed in Fig. \ref{fig:general1} presents the challenging nature of our benchmark. For example, the advanced GPT4V only attains a 67.7/100 overall score on our TinyLVLM-eHub, indicating plenty of room for improvement.
Besides the above comprehensive quantitative evaluation of LVLMs, we also present a productivity test for top-performing LVLMs with various demos. By evaluating Bard, Claude3-Opus, GeminiPro-V, and GPT4V as education assistants, document analysts, GUI navigators, and code generators, we find that while LVLMs show promise in understanding elements in simpler scenarios, they encounter difficulties in tasks requiring expert domain knowledge, intricate image-text alignment, precise localization, and complex layout generation. 


The contributions of {TinyLVLM-eHub} are summarized as follows. (1) We propose a lightweight benchmark called TinyLVLM-eHub which can thoroughly evaluate various multimodal capabilities of LVLMs with only $2.1$K image-text pairs. (2) We propose ChatGPT-based Ensemble Evaluation (CEE) as a judge to assess the model's prediction and reference answer in an open-set scenario, which aligns with Human evaluation well.
(3) Our comprehensive evaluation reveals that proprietary API-based models such as GPT4V and GemoniPro-V consistently outperform prior open-source LVLMs in various multimodal capabilities. However, they still suffer from object hallucination and encounter difficulties in tasks demanding intricate multimodal capability. 
%
We hope that our work can serve as a baseline assessment for LVLMs, and encourage further investigation on foundation multimodal models. 

\section{Related Work}\label{sec:relatedwork}
\subsection{Large Vision-Language Models} 
Large vision-language models (LVLMs) have achieved remarkable progress in various multimodal tasks. Owing to the development of open-source Large Language Models (LLM) such as LLaMA \cite{touvron2023llama} and OPT \cite{zhang2022opt}, LVLMs can utilize the knowledge from LLMs and align visual features to the text
space. For example, Flamingo \cite{alayrac2022flamingo} pioneers to insert cross-attention layers into LLMs to import visual features. To further extract effective visual prompts from images, BLIP2 \cite{li2023blip} incorporates a pre-trained visual encoder with frozen LLM by a Q-Former. Motivated by the great success of the instruction-tuning pipeline in enhancing LLMs, recent work fine-tunes LVLMs with amounts of instruction-following data. For instance, LLaVA \cite{llava} constructs 158K multimodal language-image instruction-following data to train adaption parameters and LLM. Due to the great success,  LLaVA-158K instruction following data are utilized in various LVLMs such as mPLUG-owl \cite{ye2023mplug}, LLaMA-Adapter V2 \cite{gao2023llama}, Otter \cite{li2023otter} and Otter-I \cite{li2023mimic}. Moreover, MiniGPT-4 \cite{zhu2023minigpt} develops a high-quality and well-aligned instruction dataset to train one projection layer, exhibiting many multimodal capabilities. Built upon MiniGPT-4, VPGTrans \cite{zhang2023vpgtrans} employs a technique to transfer the text encoder of a BLIP2 model to Vicuna, which reduces training costs remarkably. OF-V2 builds upon advanced LLMs and exhibits good performance on many VQA tasks. Furthermore, PandaGPT \cite{su2023pandagpt} can take multimodal inputs such as image, sound, and video simultaneously and compose their semantics naturally to perform complex tasks such as detailed image description generation. In addition, the closed-source API-based models such as Bard \cite{bard}, GeminiPro-V \cite{geminiprov}, GPT4V \cite{gpt4}, and Claude3-Opus \cite{claudeopus} distinguish themselves with their exceptional multimodal capabilities. In this work, we develop an evaluation suit to assess how well these LVLMs perform in various multimodal tasks. 

\begin{table}[]
    \centering
    \caption{The comparison between TinyLVLM-eHub and prior evaluation benchmarks. `Cap’, `MC', and `Open' indicate tested multimodal capability, multi-choice and free-form answers, respectively. We can see that TinyLVLM-eHub assess various multimodal capabilities absorbed from massive tasks with only $2.1$K test samples through GPT-based evaluation techniques. `WM', `Prefix', and `GPT' denote word-matching, prefix-based and ChatGPT-based evaluation methods, respectively.}
    \label{tab:comparison-LVLM-bench}
    \scalebox{0.95}{%
        \begin{tabular}{c|ccccc}
            \toprule
           Benchmark &
             \# Sample & \# Cap & \# Task &  Answer & Eval\\
            \cmidrule(lr){1-1}\cmidrule(lr){2-6}
            POPE \cite{li2023evaluating} & 6K & 1 & 3  & MC &WM \\
            ImageNetVC \cite{xia2023imagenetvc} & 4K & 1 & 5  & MC &Prefix\\
            OCRBench \cite{liu2023hidden} & 1K & 1 & 29  & Open & WM\\
            GVT \cite{wang2023makes} & 505K &4 &4  &MC/Open & WM\\
            MMBench \cite{liu2023mmbench} & 3K & 2 & 20  & MC & WM/GPT\\
            LVLM-eHub \cite{xu2023lvlm} & 393K &5 &42 & MC/Open &WM\\
            \cmidrule(lr){1-1}\cmidrule(lr){2-6}
            TinyLVLM-eHub & 2.1K &  6 & 42  & Open &GPT\\
            \bottomrule
        \end{tabular}%
}
\end{table}
    
\subsection{Evaluation of Large Vision-Language Models}  
Lots of research activities focus on evaluating LVLMs' capabilities, which helps understand their strengths and weaknesses and guides the further development of LVLMs. For example, Li et al.~\cite{li2023evaluating} present a systematic investigation of object hallucination of LVLMs by proposing a polling-based object probing evaluation method. Moreover, ImageNetVC \cite{xia2023imagenetvc} studies how well current LVLMs can master visual commonsense knowledge. Liu et al. \cite{liu2023hidden} comprehensively evaluate the performance of LVLMs in visual recognition with text recognition such as Optical Character Recognition (OCR). GVT \cite{wang2023makes} evaluates LVLM’s visual semantic understanding and fine-grained perception capabilities. However, these studies only evaluate specific tasks with a portion of LVLMs, lacking an overall understanding of LVLMs' capabilities. Concurrent with our work, recent benchmarks \cite{yin2023lamm, fu2023mme, liu2023mmbench, xu2023lvlm} assess LVLMs' multimodal capability by experimenting with amounts of vision-language samples. The difference is that our benchmark aims to measure LVLM's performance on various multimodal tasks with high efficiency. As shown in Table \ref{tab:comparison-LVLM-bench}, our TinyLVLM-eHub assess various multimodal capabilities absorbed from massive tasks with only $2.1$K test samples. Other than quantitative evaluation, TinyLVLM-eHub also provide various demos to compare several advanced LVLMs including Bard, GPT4V, GeminiPro-V-V, and Claude3-Opus.

\section{Tiny LVLM Evaluation Hub}

\begin{figure*}[htp]
		\centering
		\includegraphics[width=0.9\linewidth]{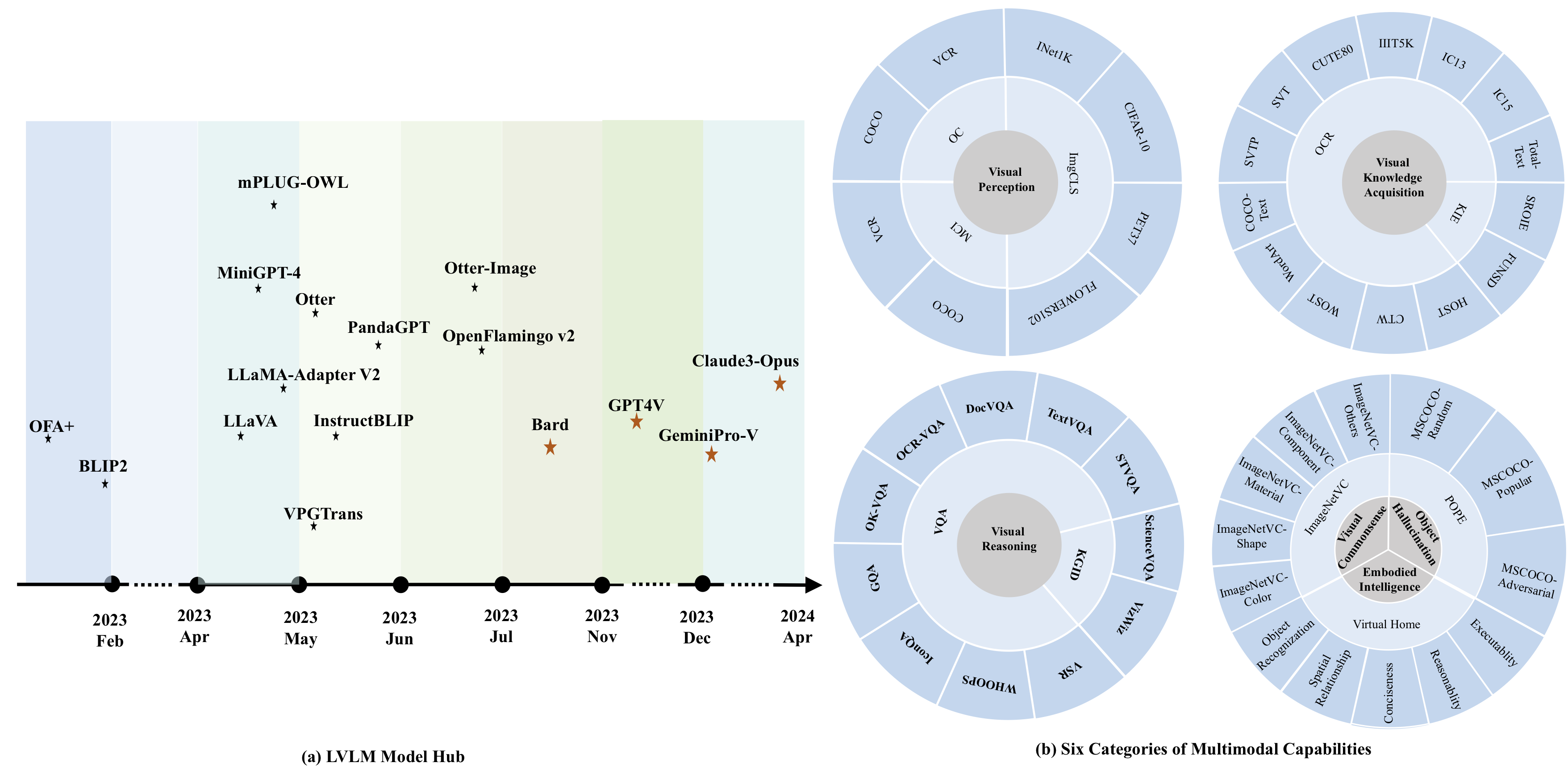}
		\caption{Visualization of TinyLVLM-eHub. (a) shows that TinyLVLM-eHub consists of 16 representative models including proprietary API-based LVLMs such as GPT4V. (b) presents six categories of capabilities tested in our TinyLVLM-eHub. }
		\label{fig:model_data}
\end{figure*}

In this section, we introduce our TinyLVLM-eHub, including an LVLM hub, the multimodal capability of interest, and the evaluation method. Compared with LVLM-eHub \cite{xu2023lvlm}, the tiny version in this work contains more LVLM models, a more lightweight sample suite, and a more accurate evaluation technique.
The overall paradigm of TinyLVLM-eHub is illustrated in Fig. \ref{fig:model_data}. 

\subsection{Model Hub}
We construct an LVLM model hub by collecting $16$ representative  LVLM models. As shown in Fig. \ref{fig:model_data}, our LVLM model hub consists of BLIP2 \cite{li2023blip}, InstructBLIP \cite{dai2023instructblip}, LLaVa \cite{liu2023visual}, LLaMA-Adapter V2 \cite{gao2023llama}, MiniGPT-4 \cite{zhu2023minigpt}, mPLUG-Owl \cite{ye2023mplug}, OF-V2 \cite{openfamingov2}, Otter \cite{li2023otter}, Otter-I \cite{li2023mimic}, PandaGPT \cite{su2023pandagpt}, VPGTrans \cite{zhang2023vpgtrans}. Moreover, we also include prior multimodal model OFA+ \cite{bai2022ofasys} and API-based LVLMs including Bard \cite{bard}, GeminiPro-V \cite{geminiprov}, GPT4V \cite{gpt4}, and Claude3-Opus \cite{claudeopus}.
The descriptions of model details have been presented in Section~\ref{sec:relatedwork}. For more information, readers are suggested to refer to their original papers. It is important to note that our access to API-based models is limited to online functionality, and we do not possess information regarding the specific configurations of these models.
As observed in LLM \cite{touvron2023llama, zhang2022opt}, the performance of an LVLM is heavily influenced by its parameter size. For comparison purposes, all the above LVLMs have parameter sizes less than 10B except for API-based models.

\subsection{Multimodal Capability}\label{sec:method-capability}

{Capability Dimension.} Following LVLM-eHub~\cite{xu2023lvlm}, we evaluate LVLMs' capability from six aspects, including visual perception, visual knowledge acquisition, visual reasoning, visual commonsense, object hallucination, and embodied intelligence. Visual perception and visual knowledge acquisition are used to detect vision ability, where visual perception like image classification is the ability to recognize the scene or objects in images while visual knowledge acquisition such as OCR needs to understand images beyond perception for knowledge acquisition. Vision Reasoning is used to assess multimodal ability, which requires a common understanding of the relationship between text as well as pictures. Moreover, the visual commonsense aims to measure the model’s comprehension of commonly shared human knowledge about generic visual concepts. Object hallucination, which is a common issue in large foundation models, measures whether the LVLM can determine the existence of objects for given images. Lastly, embodied intelligence tests the effectiveness of guiding the agent to complete a series of tasks.

{Capability Decomposition.} Fig. \ref{fig:model_data} provides an overview of the evaluation process for each multimodal capability, as demonstrated through the collection of tasks and benchmarks. This involves leveraging tasks such as Image Classification (ImgCls), Object Counting (OC), and Multi-Class Identification (MCI) to evaluate the ability of visual perception. Similarly, tasks such as Optical Character Recognition (OCR) and Key Information Extraction (KIE) are utilized to evaluate the ability of visual knowledge acquisition, while tasks of visual question answering and Knowledge-Grounded Image Description (KGID) are employed to evaluate the ability of visual reasoning. Furthermore, the dataset of ImageNetVC is used to evaluate the ability of commonsense, while the POPE \cite{li2023evaluating} pipeline is utilized to evaluate the degree of object hallucination. Finally, the benchmark of Virtual Home is utilized to evaluate the ability of embodied intelligence. The evaluation details are presented in Section~\ref{sec:exp}.

{Data Collection.} We investigate the aforementioned multimodal capabilities by collecting 42 standard text-related visual benchmarks. To create a lightweight evaluation suite, we have restricted each dataset to a total of 50 records except that Virtual Home \cite{puig2018virtualhome} in embodied intelligence has six pieces of data for efficient human annotation. Therefore, it is convenient to test various LVLMs under our TinyLVLM-eHub. As a precautionary measure, we have filtered out images that contain human portraits, vulgar content, and medical organs. This ensures that API-based LVLMs can produce results without encountering any warning messages. 

\subsection{Evaluation Method}\label{sec:evaluation methods}
\begin{figure*}[htp]
		\centering
		\includegraphics[width=1\linewidth]{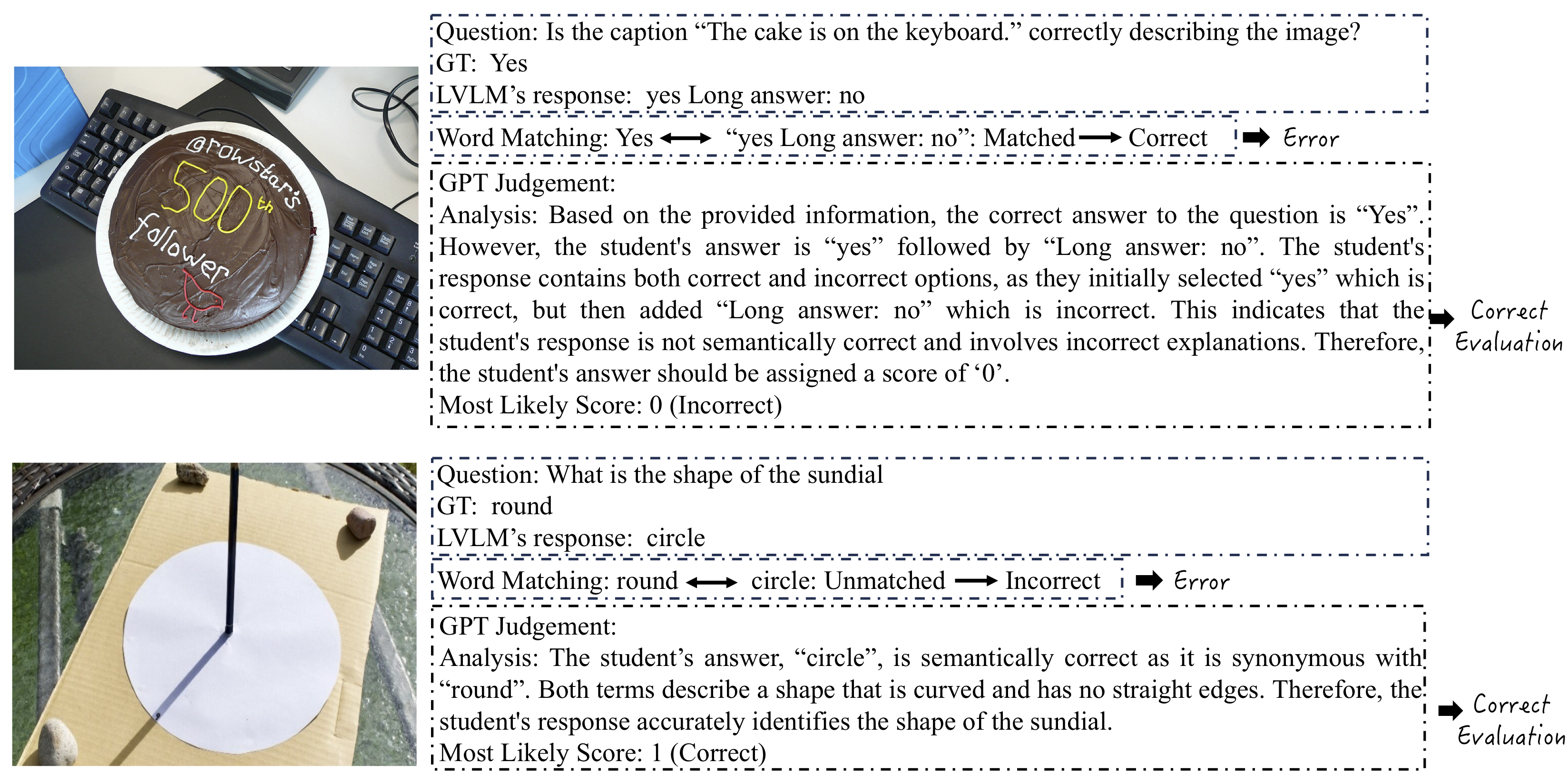}
		\caption{Two evaluation cases where word matching fails but CEE succeeds. In the upper one, the model's output is self-contradictory and includes all possible options where the word matching evaluation can \emph{cheat}. For the bottom sample, the model generates a different expression (\emph{i.e.}, paraphrasing in general) from the ground truth. While they essentially have the same meaning, word matching definitely fails.}
		\label{fig:cee_better}
\end{figure*}
\begin{figure*}[htp]
		\centering
		\includegraphics[width=1\linewidth]{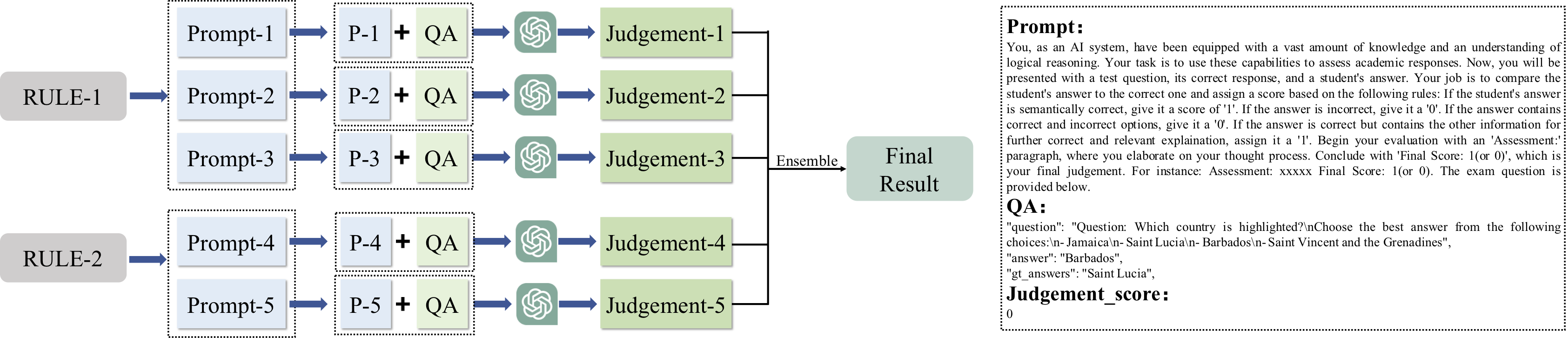}
		\caption{Illustration of our proposed evaluation methods. 
		}
		\label{fig:evalation-methods}
\end{figure*}
We use zero-shot evaluation to assess LVLMs. To this end, LVLM-eHub \cite{xu2023lvlm} assesses the predicted answer and reference answer by \emph{word matching} (\emph{i.e.}, the prediction is correct as long as it exists in the reference answer). However, simple word matching is not effective in comparing answer pairs as illustrated in Fig.~\ref{fig:cee_better}. Although recent works \cite{liu2023mmbench, fu2023mme} assess the predicted answer and the reference answer by constraining the model's output to be fixed forms (e.g. Yes/No or A/B/C/D), they fail to evaluate LVLMs' capability in the open-set setting.

To tackle the above drawbacks, we introduce a new evaluation metric called ChatGPT Ensemble Evaluation (CEE) which consists of a diverse prompt generation and an ensemble voting, as shown in Fig.~\ref{fig:evalation-methods}. Specifically, CEE first customizes two rules for prompt generations. For each rule, several prompts are generated by GPT-4. Given each tuple of (prompt, question, reference answer, predicted answer), ChatGPT is employed to determine whether the predicted answer is correct, which is a promising assessment \cite{zheng2023judging}. Due to a good understanding of the texts of ChatGPT, CEE allows for free-form predicted answers. Finally, CEE votes for the final answer by collecting all judgments from ChatGPT. With diverse prompts ensemble, CEE can be robust and accurate in evaluating the performance of LVLMs. An example of prompt and ChatGPT's judgment is given in Fig.~\ref{fig:evalation-methods}. We show that our CEE can align with Human’s evaluation better than the word matching by experiments in Section~\ref{sec:exp-ablation-metric}.

\section{Experiment and Analysis}\label{sec:exp}
This section focuses on evaluating the performance of LVLMs in various areas such as visual perception, visual knowledge acquisition, visual reasoning, visual commonsense understanding, visual object hallucination, and embodied intelligence, as detailed in Sections \ref{sec:exp-eval-results}. Additionally, we provide an ablation study for the benchmark in Section \ref{sec:exp-ablation-metric}. Lastly, we present multimodal applications of top-performing LVLMs by productivity test in Section~\ref{sec:exp-bard-demo}. The summarized results are reported in Fig. \ref{fig:general1}.

\subsection{Evaluation Results on Multimodal Capability} \label{sec:exp-eval-results}
\subsubsection{Results on Visual Perception.}\label{sec:exp-perception}

We evaluate the ability of visual perception through tasks of ImgCls, MCI, and OC. The results are reported in Table \ref{tab:oc_mci_results_gpt}. It can be seen that GPT4V outperforms other LVLMs by a large margin on average. OFA+ achieves the worst performance in visual perception. It indicates that the fashion of the clip encoder and large language model used in current LVLMs is superior to the previous model architecture in OFA+. 
In addition, API-based LVLMs exhibit evident advantages over other LVLMs on the task of object counting and multi-class identification.  
However, the performance of these models on CIFAR10 lags behind open-source LVLMs. It may be caused by the low resolution of images in CIFAR10.

\begin{table*}[t]
\centering
\small
\caption{Evaluation results of visual perception capability of LVLMs on tasks of Image Classification (ImgCls), Object Counting (OC), and Multi-class Identification (MCI). The accuracy is used to measure the performance of all datasets. The {best} result is in {bold}. 
The average score means the accuracy over all datasets on average.}
\label{tab:oc_mci_results_gpt}
\resizebox{\textwidth}{!}{%
\begin{tabular}{c|cccc |cc|cc|c}
\toprule
\multirow{2}{*}{Model}
&\multicolumn{4}{c|}{ImgCls} & \multicolumn{2}{c|}{OC~\cite{gvt}} & \multicolumn{2}{c|}{MCI~\cite{gvt}}  & \multirow{2}{*}{Avg. Score} \\
 & INet1K \cite{imagenet}& CIFAR10 \cite{Krizhevsky2009LearningML} & Pets37 \cite{parkhi12a} & FLowers102
 ~\cite{Nilsback08} & COCO~\cite{chen2015microsoft} & VCR~\cite{zellers2019recognition} & COCO & VCR & 
\\
\cmidrule(lr){1-1}\cmidrule(lr){2-5}\cmidrule(lr){6-7}\cmidrule(lr){8-9}\cmidrule(lr){10-10}
OFA+& 14& 42& 12& 2& 62& 74& 42& 78& 40.8 \\
\cmidrule(lr){1-1}\cmidrule(lr){2-5}\cmidrule(lr){6-7}\cmidrule(lr){8-9}\cmidrule(lr){10-10}
BLIP2 & {68} & 86 & 40 & 38 & 48 & 20 & {88} & {88} & {59.5} \\ 
InstructBLIP & {64} & {90} & 30 & {46} & {62} & {30} & {78} & 66 & {58.3} \\ 
LA-V2 & 54 & 74 & 40 & 34 & 48 & {30} & 70 & 64 & 51.8 \\ 
LLaVA & 44 & 86 & 20 & 18 & 34 & {30} & 48 & 46 & 40.8 \\ 
MiniGPT-4 & 52 & 64 & 28 & 36 & 38 & 18 & 48 & 46 & 41.3 \\ 
mPLUG-Owl & 42 & 72 & 56 & 38 & 32 & 18 & 56 & 42 & 44.5 \\ 
OF-V2 & 42 & 82 & {60} & {42} & 6 & 12 & 50 & 46 & 42.5 \\ 
Otter & 22 & 52 & 20 & 14 & 46 & {30} & 64 & 44 & 36.5 \\ 
Otter-I & 50 & 82 & 8 & 10 & 56 & {30} & 48 & 44 & 41.0 \\ 
PandaGPT & 44 & 76 & 8 & 2 & 36 & {34} & 70 & 60 & 41.3 \\ 
VPGTrans & 54 & {92} & 30 & 28 & 38 & 18 & 32 & 52 & 43.0 \\ 
\cmidrule(lr){1-1}\cmidrule(lr){2-5}\cmidrule(lr){6-7}\cmidrule(lr){8-9}\cmidrule(lr){10-10}
Bard & 58 & 46 & {58} & 40 & {60} & 22 & 72 & {80} & 54.5 \\
GeminiPro-V & 38 & 60 & 80 & 52 & 62 & 90 & 36 & 88 & 63.3 \\
GPT4V & 50 & 54 & 84 & 76 & 62 & 94 & 30 & 88 & \textbf{67.2} \\
Claude3-Opus & 54 & 26 & 84 & 80 & 28 & 82 & 12 & 78 & 55.5 \\
\bottomrule
\end{tabular}%
}
\end{table*}

\begin{table*}[t]
\centering
\small
\caption{ Comparison of Zero-shot Performance for Large-scale Vision and Language Models (LVLMs) on OCR and KIE Tasks.}
\label{tab:vka_results}
\resizebox{\textwidth}{!}{%
\begin{tabular}{c|ccc ccc ccc ccc| cc | c}
\toprule
\multirow{2}{*}{Model}
&\multicolumn{12}{c|}{OCR} & \multicolumn{2}{c|}{KIE} & \multirow{2}{*}{Avg. Score} \\
 & IIIT5K~\cite{ittt5k} & IC13~\cite{ic13} & IC15~\cite{ic15} & Total-Text~\cite{total-text} & CUTE80~\cite{cute80} & SVT~\cite{svt} & SVTP~\cite{svtp} & COCO-Text~\cite{coco-text} & WordArt~\cite{wordart} & CTW~\cite{ctw} & HOST~\cite{ost} & WOST~\cite{ost} & SROIE~\cite{sroie} & FUNSD~\cite{funsd} \\
\cmidrule(lr){1-1}\cmidrule(lr){2-13}\cmidrule(lr){14-15}\cmidrule(lr){16-16} 
OFA+& 0& 2& 0& 4& 0& 2& 0& 2& 0& 0& 0& 0& 2& 0& 0.9 \\
\cmidrule(lr){1-1}\cmidrule(lr){2-13}\cmidrule(lr){14-15}\cmidrule(lr){16-16} 
BLIP2 & {68} & {86} & 52 & 58 & 72 & {76} & {72} & {48} & 40 & 54 & {50} & {60} & 0.0 & 4.0 & 52.9 \\
InstructBLIP & {82} & {86} & {66} & {62} & {78} & {76} & {74} & {56} & {48} & {56} & {54} & {56} & 0.0 & 4.0 & {57.0} \\
LA-V2 & 52 & 20 & 28 & 28 & 28 & 20 & 26 & 20 & 36 & 20 & 16 & 6.0 & 4.0 & {56} & 25.7 \\
LLaVA & 14 & 8.0 & 14 & 12 & 24 & 2.0 & 10 & 14 & 26 & 12 & 14 & 8.0 & 2.0 & 42 & 14.4 \\
MiniGPT-4 & 20 & 18 & 8.0 & 6.0 & 16 & 10 & 6.0 & 12 & 20 & 6.0 & 6.0 & 4.0 & 2.0 & 20 & 11.0 \\
mPLUG-Owl & 22 & {4.0} & 20 & 14 & 24 & 4.0 & 8.0 & 8.0 & 24 & 8.0 & 6.0 & 2.0 & 0.0 & 16 & 11.4 \\
OF-V2 & 28 & 18 & 24 & 20 & 22 & 10 & 22 & 18 & 28 & 16 & 14 & 6.0 & 0.0 & 28 & 18.1 \\
Otter & 4.0 & 6.0 & 8.0 & 10 & 6.0 & 6.0 & 4.0 & 6.0 & 8.0 & 4.0 & 2.0 & 4.0 & 0.0 & 26 & 6.7 \\
Otter-I & 18 & 6.0 & 18 & 24 & 16 & 12 & 12 & 16 & 26 & 8.0 & 14 & 2.0 & 2.0 & 38 & 15.1 \\
PandaGPT & 2.0 & 0.0 & 0.0 & 4.0 & 4.0 & 0.0 & 0.0 & 4.0 & 0.0 & 2.0 & 0.0 & 0.0 & 0.0 & 22 & 2.7 \\
VPGTrans & 50 & 80 & 36 & 56 & 42 & 64 & 62 & 32 & 42 & 44 & 46 & 42 & 0.0 & 22 & 44.1 \\
\cmidrule(lr){1-1}\cmidrule(lr){2-13}\cmidrule(lr){14-15}\cmidrule(lr){16-16}
Bard & {78} & {84} & {60} & {60} & {76} & {74} & 66 & 42 & {54} & {64} & 46 & 52 & {42} & {50} & {60.6} \\
GeminiPro-V & 86 & 90 & 52 & 60 & 72 & 82 & 60 & 32 & 68 & 48 & 46 & 40 & 60 & 46 & 60.1 \\
GPT4V & 78& 70& 42& 54& 72& 44& 32& 38& 74& 58& 28& 38& 78& 86& 56.6 \\
Claude3-Opus & 86& 86& 70& 60& 84& 80& 74& 32& 66& 60& 48& 52& 74& 58& \textbf{66.4}\\
\bottomrule
\end{tabular}%
}
\end{table*}

\subsubsection{Results on Visual Knowledge Acquisition}
We evaluate the ability in visual knowledge acquisition through tasks of OCR and KIE. The results are shown in Table~\ref{tab:vka_results}. We can see that API-based models still outperform other LVLMs by a large margin. Note that these models achieve remarkable success on the KIE task compared with other LVLMs, implying that API-based models can recognize characters and aggregate useful information in the image. From Table~\ref{tab:vka_results}, Claude3-Opus consistently achieves strong performance on various OCR and KIE tasks, indicating that Claude3-Opus is good at acquiring detailed information in the image.

\begin{table*}[t]
\centering
\small
\caption{ Comparison of Zero-shot Performance for LVLM Models on VQA and KGID Tasks. In these experiments, top-1 accuracy is employed for evaluation.}
\label{tab:vr_results}
\resizebox{\textwidth}{!}{%
\begin{tabular}{c|ccc ccc ccc| cc | c}
\toprule
\multirow{2}{*}{Model}
&\multicolumn{9}{c|}{VQA} & \multicolumn{2}{c|}{KGID} & \multirow{2}{*}{Avg. Score} \\
 & DocVQA~\cite{docvqa} & TextVQA~\cite{textvqa} & STVQA~\cite{stvqa} & OCR-VQA~\cite{ocr-vqa} & OKVQA~\cite{okvqa} & GQA~\cite{gqa}  & IconQA~\cite{iconqa} & VSR~\cite{vsr} & WHOOPS~\cite{whoops} & ScienceQA~~\cite{scienceqa} & VizWiz~\cite{vizwiz}  \\
\cmidrule(lr){1-1}\cmidrule(lr){2-10}\cmidrule(lr){11-12}\cmidrule(lr){13-13}
OFA+& 2& 0& 16& 20& 22& 50& 16& 44& 8& 4& 28& 19.1 \\ 
\cmidrule(lr){1-1}\cmidrule(lr){2-10}\cmidrule(lr){11-12}\cmidrule(lr){13-13} 
BLIP2 & 6.0 & 36 & 40 & 52 & 52 & 36 & 46 & {66} & {56} & {66} & {66} & 47.5 \\
InstructBLIP & 10 & 40 & 52 & {76} & {66} & {58} & 42 & 54 & 42 & 48 & {78} & {51.5} \\
LA-V2 & 20 & {54} & {58} & 50 & 58 & {44} & 44 & 52 & 40 & 56 & 54 & 48.2 \\
LLaVA & 8.0 & 34 & 42 & 34 & 34 & {44} & 40 & 52 & 30 & 54 & 64 & 39.6 \\
MiniGPT-4 & 12 & 34 & 30 & 34 & 36 & 20 & 32 & 48 & 22 & 6.0 & 38 & 28.4 \\
mPLUG-Owl & 2.0 & 28 & 26 & 18 & 16 & 20 & 22 & 46 & 12 & 10 & 26 & 20.5 \\
OF-V2 & 8.0 & 34 & 52 & 44 & 34 & 40 & {48} & 58 & 32 & 48 & 58 & 41.5 \\
Otter & 10 & 24 & 30 & 28 & 54 & 20 & 34 & 24 & 12 & 34 & 46 & 28.7 \\
Otter-I & 14 & 40 & 46 & 34 & 50 & 44 & 36 & 56 & 20 & 48 & 54 & 40.2 \\
PandaGPT & 10 & 16 & 24 & 30 & 48 & 38 & 34 & 60 & 14 & 50 & 42 & 33.3 \\
VPGTrans & {22} & 38 & 42 & 32 & 36 & 34 & 32 & 40 & 36 & 12 & 48 & 33.8 \\
\cmidrule(lr){1-1}\cmidrule(lr){2-10}\cmidrule(lr){11-12}\cmidrule(lr){13-13}
Bard & {48} & {60} & {72} & {80} & {68} & 40 & {62} & {82} & {42} & {68} & 62 & {62.2} \\
GeminiPro-V & 56& 66& 78& 62& 36& 38& 58& 62& 40& 74& 50& 56.4 \\
GPT4V & 76& 84& 66& 72& 58& 36& 76& 72& 30& 74& 66& \textbf{64.5}\\
Claude3-Opus & 86& 80& 64& 64& 50& 24& 52& 52& 42& 82& 68& 60.4\\
\bottomrule
\end{tabular}%
}
\end{table*}

\subsubsection{Results on Visual Reasoning.}
We evaluate the ability in visual reasoning on tasks of VQA and KGID. 
The results are shown in Table~\ref{tab:vr_results}. We draw several conclusions. First, GPT4V  achieves the best performance in visual reasoning compared with other LVLMs. It shows that GPT4V has a comprehensive understanding of images and related texts. Second, API=based models obtain less competitive performance than BLIP2 on WHOOPS whose VQA samples are created by breaking commonsense, implying that the commonsense understanding can be further improved. Third, API-based models also have a good understanding of science knowledge because they achieve good performance in ScienceQA which denotes the question-answering pairs with visual inputs in ScienceQA \cite{scienceqa}.

\begin{table}[t]
\centering
\small
\caption{ Comparisons of Zero-shot visual commonsense Performance for LVLM Models on ImageNetVC datasets. Top-1 accuracy is employed for the evaluation.}
\label{tab: visual commonsense}
\resizebox{.5\textwidth}{!}{%
\begin{tabular}{c|ccc cc | c}
\toprule
\multirow{2}{*}{Model}
&\multicolumn{5}{c|}{ImageNetVC~\cite{xia2023imagenetvc}} & \multirow{2}{*}{Avg. Score} \\
 & Color & Shape & Material & Component & Others  \\
\cmidrule(lr){1-1}\cmidrule(lr){2-6} \cmidrule(lr){7-7} 
OFA+& 54& 28& 38& 48& 57& 45.2 \\ 
\cmidrule(lr){1-1}\cmidrule(lr){2-6} \cmidrule(lr){7-7} 
BLIP2 & 32 & 16 & 36 & {76} & 66 & 45.2 \\
InstructBLIP & {52} & {58} & {64} & {76} & {76} & {65.2} \\
LA-V2 & 42 & 38 & {62} & {76} & {72} & {58.0} \\
LLaVA & 42 & 38 & 50 & 50 & 54 & 46.8 \\
MiniGPT-4 & 30 & 28 & 36 & 50 & 32 & 35.2 \\
mPLUG-Owl & 14 & 16 & 34 & 26 & 28 & 23.6 \\
OF-V2 & 44 & 32 & 48 & 56 & 48 & 45.6 \\
Otter & 36 & 30 & 44 & 52 & 64 & 45.2 \\
Otter-I & 46 & 40 & 54 & 60 & 64 & 52.8 \\
PandaGPT & {48} & 34 & 48 & 64 & 58 & 50.4 \\
VPGTrans & 36 & {48} & 46 & 70 & 48 & 49.6 \\
\cmidrule(lr){1-1}\cmidrule(lr){2-6}\cmidrule(lr){7-7}
Bard & 40 & 44 & 52 & {82} & 72 & {58.0} \\
GeminiPro-V& 56& 62& 50& 82& 60& 62.0 \\ 
GPT4V& 54& 66& 58& 90& 82& 70.0 \\ 
Claude3-Opus& 52& 68& 70& 86& 80& \textbf{71.2} \\ 
\bottomrule
\end{tabular}%
}
\end{table}

\begin{table*}[t]
\centering
\small
\caption{ Detailed evaluation results of the zero-shot performance of LVLMs on MSCOCO using POPE evaluation pipeline \cite{li2023evaluating}, where Acc represents the accuracy of prediction; Prec represents how many of the predicted positive samples are true positive samples; Recall represents how many of all true positive samples are correctly identified; and Yes represents the probability that the model outputs a yes answer. The average score is calculated based on the metric of accuracy.}
\label{tab:pope_results}
\resizebox{\textwidth}{!}{%
\begin{tabular}{c|cccc| cccc| cccc|  c}
\toprule
\multirow{2}{*}{Model}
&\multicolumn{4}{c|}{MSCOCO-Random \cite{li2023evaluating}} &\multicolumn{4}{c|}{MSCOCO-Popular \cite{li2023evaluating}} &\multicolumn{4}{c|}{MSCOCO-Adversarial \cite{li2023evaluating}} & \multirow{2}{*}{Avg. Score} \\
 & Acc & Prec & Recall & Yes  & Acc & Prec & Recall & Yes & Acc & Prec & Recall & Yes  \\
\cmidrule(lr){1-1}\cmidrule(lr){2-5} \cmidrule(lr){6-9} \cmidrule(lr){10-13}\cmidrule(lr){14-14} 
OFA+& 80& 100 & 75 & 44 & 80& 100 & 84 & 42 & 76& 100 & 84 & 42 & 78.7 \\
\cmidrule(lr){1-1}\cmidrule(lr){2-5} \cmidrule(lr){6-9} \cmidrule(lr){10-13}\cmidrule(lr){14-14} 
BLIP2 & 72 & {100} & 52 & 30 & {86} & 87 & 86 & 48 & {90} & {95} & 84 & 44 & {82.7} \\
InstructBLIP & {82} & {100} & 81 & 39 & {92} & 92 & {92} & 49 & {92} & 92 & {92} & 49 & \textbf{88.7} \\
LA-V2 & 64 & 59 & 76 & 74 & 46 & 43 & 72 & 84 & 46 & 45 & 80 & 88 & 52.0 \\
LLaVA & 54 & 54 & {93} & {100} & 46 & 46 & 92 & {100} & 40 & 40 & 80 & {100} & 46.7 \\
MiniGPT-4 & 65 & 71 & 71 & 60 & 63 & 71 & 74 & 56 & 46 & 44 & 53 & 61 & 58.0 \\
mPLUG-Owl & 68 & 63 & {100} & 86 & 66 & 61 & 95 & 88 & 59 & 60 & 86 & 81 & 64.3 \\
OF-V2 & 54 & 55 & 93 & {98} & 48 & 49 & {96} & 98 & 50 & 50 & {96} & {96} & 50.7 \\
Otter & 47 & 44 & 71 & 92 & 42 & 42 & 84 & {100} & 44 & 44 & 84 & {96} & 44.3 \\
Otter-I & {76} & {88} & 76 & 50 & 68 & 69 & 88 & 64 & 66 & 66 & {92} & 70 & 70.0 \\
PandaGPT & 58 & 56 & {93} & 96 & 50 & 48 & {92} & {96} & 48 & 46 & 88 & {96} & 52.0 \\
VPGTrans & 67 & 92 & 46 & 28 & 80 & {94} & 65 & 35 & 79 & 88 & 65 & 36 & 75.3 \\
\cmidrule(lr){1-1}\cmidrule(lr){2-5}\cmidrule(lr){6-9}\cmidrule(lr){10-13}\cmidrule(lr){14-14}
Bard & 63 & {100} & 36 & 18 & 70 & {100} & 40 & 18 & 69 & {100} & 43 & 19 & 67.3 \\
GeminiPro-V& 70&  94&  52&  32& 86&  87&  84&  48& 84&  90&  76& 42 & 80.0 \\ 
GPT4V& 70&  89&  55&  36& 86& 84 &  88& 52 & 74& 70 &  84&  60& 76.7 \\ 
Claude3-Opus& 68&  84&  55&  38& 80&  82&  76& 46 & 70&  69&  72&  52& 72.7 \\ 
\bottomrule
\end{tabular}%
}
\end{table*}

\begin{table}[h]
\centering
\small
\caption{Generated planning quality evaluation on embodied tasks. 10 participants are involved in the user study for evaluation. The evaluation comprises five dimensions with scores ranging from $1$ to $5$, including object recognition (OR), spatial relationship (SR), level of conciseness (Con.), reasonability (Rea.), and executability (Exe.) of the planning. The final score for each dimension is averaged over all participants and normalized by $(\cdot )/5 \times 100\%$. Bard$^*$ means that only samples without describing humans are included. We see that Bard exhibits good planning ability for embodied application.}
\label{tab:embodied_tasks_eval}
\resizebox{.48\textwidth}{!}{%
\begin{tabular}{c|ccc cc | c}
\toprule
\multirow{2}{*}{Model}
&\multicolumn{5}{c|}{Virtual Home~\cite{puig2018virtualhome}} & \multirow{2}{*}{Avg. Score} \\
 & OR &  SR & Con. & Rea. & Exe.  \\
\cmidrule(lr){1-1}\cmidrule(lr){2-6} \cmidrule(lr){7-7} 
OFA+ &22.5& 21.7& 49.2& 20.8& 20.0& 26.8\\
\cmidrule(lr){1-1}\cmidrule(lr){2-6} \cmidrule(lr){7-7} 
BLIP2 & 40.6 & 33.6 & 65.0 & 55.6 & 57.6 & 50.4 \\
InstructBLIP & 61.6 & 55.6 & 49.6 & 64.0 & 62.0 & 58.6 \\
LA-V2 & 76.2 & 74.2 & 59.2 & 80.8 & 81.6 & 74.4 \\
LLaVA & {77.6} & 72.2 & 37.2 & 74.0 & 76.4 & 67.4 \\
MiniGPT-4 & 74.0 & 69.4 & 32.4 & 70.8 & 62.2 & 61.8 \\
mPLUG-Owl & 68.4 & 64.4 & 29.6 & 68.8 & 70.8 & 60.4 \\
OF-V2 & 23.2 & 24.2 & {77.2} & 37.0 & 35.8 & 39.4 \\
Otter & 67.6 & 62.0 & 37.2 & 61.4 & 62.4 & 58.2 \\
Otter-I & {81.0} & {85.0} & 57.8 & 76.4 & 80.0 & 76.0 \\
PandarGPT & 74.8 & 74.6 & 65.8 & {89.4} & {89.4} & {78.8} \\
VPGTrans & 68.6 & 64.4 & 35.2 & 67.0 & 67.0 & 60.4 \\
\cmidrule(lr){1-1}\cmidrule(lr){2-6} \cmidrule(lr){7-7} 
Bard$^*$ & 73.0 & {79.8} & {79.0} & {94.2} & {91.8} & \textbf{83.6} \\
GeminiPro-V& 88.3& 86.7& 31.7& 87.5& 89.2& 76.7 \\ 
GPT4V& 52& 68& 70& 86& 80& 71.2 \\ 
Claude3-Opus& 54& 66& 58& 90& 82& 70.0 \\ 
\bottomrule
\end{tabular}%
}
\end{table}

\subsubsection{Results on Visual Commonsense.}
We perform the visual commonsense evaluation on the ImageNetVC dataset which asks the common concepts of visual input, such as color, shape, material, component, and others.
Table~\ref{tab: visual commonsense} presents the results. We see that GPT4v and Claude3-Opus present good performance on this task. In particular, Claude3-Opus exhibits a good understanding of all aspects while GPTV lags in object materials on ImageNetVC. Bard has a good understanding of commonsense but leaves room for improvement. Specifically, we can see from Table \ref{tab: visual commonsense} that Bard does not well comprehend the commonsense related to color, shape, and material compared with InstructBLIP \cite{dai2023instructblip}.

\subsubsection{Results on Object Hallucination.} 
We test the degree of object hallucination of Bard on MSCOCO under the POPE framework \cite{li2023evaluating} which asks YES/NO questions about the existence of objects given an image. We report results in terms of accuracy, precision, recall, and Yes (the ratio of answering Yes). The results are shown in Table \ref{tab:pope_results}. We can see that API-based models achieve less satisfactory performance than the other 2 LVLM models, including InstructBLIP, and BLIP, showing that they still suffer from object hallucination. By comparing the results of precision, recall, and yes, we find that Bard tends to stick in the mud and often answers `no' even when the object indeed exists in the image. Such object hallucination is different from the type of other LVLMs which tends to answer `yes' even when the object does not exist in the image. To our surprise, OFA+ achieves much better performance on object hallucination than many LVLMs. This indicates that the hallucination problem might come from the LLM decoder in LVLMs. Our experiment reveals that the object hallucination issue of LVLMs still remains a challenging problem.

\subsubsection{Results on Embodied Intelligence.}\label{sec:exp-embodied}
We present the evaluation results on embodied intelligence. Similar to LVLM-eHub \cite{xu2023lvlm}, we conducted a user study involving 10 participants to assess the effectiveness of planning outputs. The study comprises 6 household scenarios from VirtualHome~\cite{puig2018virtualhome}.  The results are reported in Table~\ref{tab:embodied_tasks_eval}.  Given that Bard is specifically designed for images devoid of human presence, we present evaluation results for Bard on test splits both with and without human subjects. Bard (w/o human) garners the highest average score across five key dimensions and exhibits unmatched performance in terms of reasonability and executability.  However, its proficiency in Object Recognition fails to rank among the top three, highlighting limitations in its fine-grained visual detection capability within the embodied domain. Moreover, acknowledging the frequent occurrence of human presence, and even direct human interaction in daily embodied tasks, it's evident that Bard has considerable room for improvement. In addition, we see that  GminiPro-V achieves the second-best performance but attains a low score in conciseness. Striking a balance between maintaining human preference and ensuring task efficiency presents a substantial development frontier for API-based models in the embodied domain tasks.

\subsection{Ablation Study}\label{sec:exp-ablation-metric}

\subsubsection{The representativeness of TinyLVLM-eHub} Note that each dataset of TinyLVLM-eHub consists of $50$ samples from the original LVLM-eHub \cite{xu2023lvlm} except for tasks in embodied intelligence. We verify the representativeness of TinyLVLM-eHub by calculating Pearson's correlation between score sequences of LVLMs existing in both two benchmarks (BLIP2, LA-V2, LLaVA, MiniGPT-4, mPLUG-Owl, and Otter). We collect Pearson's correlations on overall score and five multimodal capabilities including visual perception, reasoning, knowledge acquisition, commonsense, and hallucination. 
As shown in Table \ref{tab:corr_lvlmehubs}, the overall score of LVLMs in TinyLVLM-eHub is highly correlated (0.98) with that in LVLM-eHub. The correlation score on each capability is more than 0.8. These results validate the efficacy of the TinyLVLM-eHub as a proxy for the comprehensive LVLM-eHub, thus ensuring that the lightweight benchmark still provides a robust indication of model performance. The lightweight nature of TinyLVLM-eHub will be accessible to a wider range of participants.

\begin{table}[h]
\centering
\small
\caption{The Pearson’s correlations between the score sequences of LVLMs existing both in TinyLVLM-eHub and LVLM-eHub are calculated. The results are obtained on the overall score and scores on five multimodal capabilities.
}
\label{tab:corr_lvlmehubs}
\resizebox{.48\textwidth}{!}{%
\begin{tabular}{c|ccc cc c }
\toprule
 & Overall Score & Perception & Knowledge & Reasoning & Commonsense & Hallucination  \\
\cmidrule(lr){1-1}\cmidrule(lr){2-7} 
Pearson's $r$ &0.98 & 0.86 & 0.99 &0.98 &0.87 &0.80\\
\bottomrule
\end{tabular}%
}
\end{table}

\subsubsection{Compared with word-matching evaluation method}
We ablate evaluation methods of word matching and CEE in terms of agreement with human evaluation in Table~\ref{tab:gpt_vs_hasword}. Among all datasets studied in this work, we manually select 5 representative and diverse ones, \emph{namely} IC15, ImageNetVC shape, MSCOCO POPE adversarial, VCR MCI, and VSR, to conduct the human evaluation. As illustrated in Table~\ref{tab:gpt_vs_hasword}, 11 out of 12 models show noticeably better agreement (\emph{i.e.}, accuracy averaged over all samples of 5 datasets) of CEE than word matching with human annotation, while both methods enjoy a high agreement of greater than 80\%.

Bard is the only exception where CEE performs worse than word matching. LLMs are well known for being capable of generating long, complex, and coherent text. Bard is much more talkative than others and hence more likely inclined to output verbose responses. Therefore, from another perspective, Bard is also more competent in fabricating unfaithful and/or nonfactual statements or reasoning.  Besides, due to its close source, we have no explicit or guaranteed way to control its generation length as we can do to open source LVLMs (\emph{i.e.}, \texttt{max\_new\_tokens}). Empirically, while Bard indeed can follow the specified answer format in the prompt, it continues generation after formatted answers and usually ends with irrelevant messages. Based on the observations above, we hypothesize that Bard's coherent but talkative responses with possible hallucinations could hurt CEE, especially when ground-truth answers of those 5 chosen datasets are all succinct, close-set, and definite. 

\begin{table*}[t]
\centering
\small
\caption{The comparison of the alignment with human evaluation between the word matching approach~\cite{xu2023lvlm} and our ChatGPT Ensemble Evaluation (CEE). Higher alignment indicates more consistent evaluation with human annotation. We see that CEE achieves higher alignment on all LVLMs except for Bard and aligns generally better with human evaluation when more prompts are employed for ensemble evaluation.}
\label{tab:gpt_vs_hasword}
\resizebox{\textwidth}{!}{%
    \begin{tabular}{c|ccc ccc ccc ccc| c}
    \toprule
     Evaluation & BLIP2 & InstructBLIP & LA-V2 & LLaVA & MiniGPT-4 & mPLUG-Owl & OF-V2  & Otter &Otter-I & PandaGPT & VPGTrans & Bard & Avg \\
    \cmidrule(lr){1-1}\cmidrule(lr){2-13}\cmidrule(lr){14-14}
    Word Matching \cite{xu2023lvlm} & 85.0 & 86.0 & 90.0 & 85.2 & 85.6 & 87.6 & 83.2 & 80.8 & 92.0 & 82.4 & 85.6 & 92.0 &86.3 \\
    CEE (3 prompts) & 87.1 & 89.6 &88.6&89.6&86.4&88&88.2&83.2&92.8&88.2&88.6&88.4 &88.2 \\
 CEE (4 prompts) & 88.0& 90.4& 88.8& 88.8& 86.8& 88.4& 89.6& 83.6& 92.8& 88.0& 88.0& 85.8 &88.3\\
  \cmidrule(lr){1-1}\cmidrule(lr){2-13}\cmidrule(lr){14-14}
    CEE (5 prompts ours) & {89.2} & {90.0} & {90.8} & {89.6} & {87.6} & {90.0} & {90.8} & {82.4} & {92.8} & {88.0} & {87.6} & 86.4 &88.9 \\
    \bottomrule
    \end{tabular}%
}
\end{table*}

\subsubsection{The effect of the number of prompts used in CEE}
We investigate the effect of the number of prompts used in the proposed CEE. Specifically, we use CEE with 3/4/5 prompts ensemble to evaluate LVLMs and report the alignment with human evaluation. As shown in Table \ref{tab:gpt_vs_hasword}, CEE generally aligns better with human evaluation when more prompts are employed for ensemble evaluation.

\begin{figure}[t]
		\centering
		\includegraphics[width=1.0\linewidth]{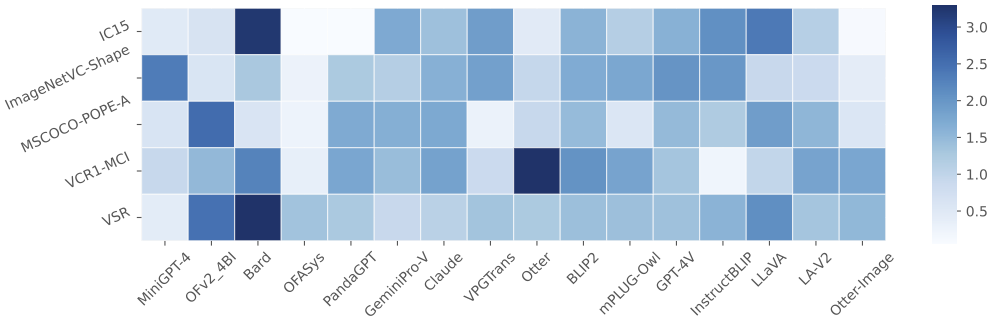}
		\caption{ The sensitivity of ChatGPT Ensemble Evaluation (CEE) to random noise by visualizing the heatmap of score variances across $16$ LVLMs. These results are based on evaluations conducted on $5$ datasets with three independent evaluations each. Our findings indicate that CEE evaluation results demonstrate robustness to random noise, as evidenced by the small variance observed across multiple evaluation processes.}
		\label{fig:llm-bias}
\end{figure}

\begin{figure}[t]
		\centering
		\includegraphics[width=1.0\linewidth]{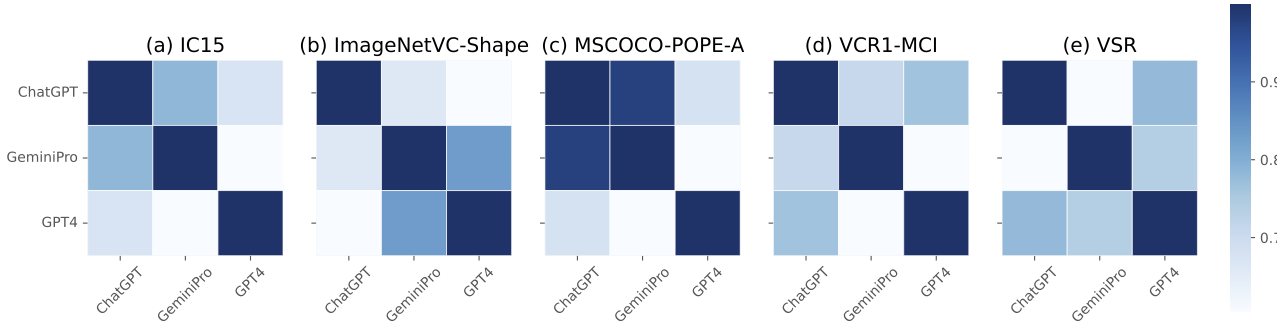}
		\caption{The sensitivity of CEE to different LLMs. We report the correlation between overall scores of $16$ LVLMs using the proposed CEE evaluation with different LLMs as judges. The LLM judges comprise ChatGPT, GPT4, and language-only GeminiPro. The overall score sequences obtained by different LLMs are highly correlated with each other.}
		\label{fig:llm-noise}
\end{figure}

\begin{figure*}[h]
		\centering
		\includegraphics[width=1.0\linewidth]{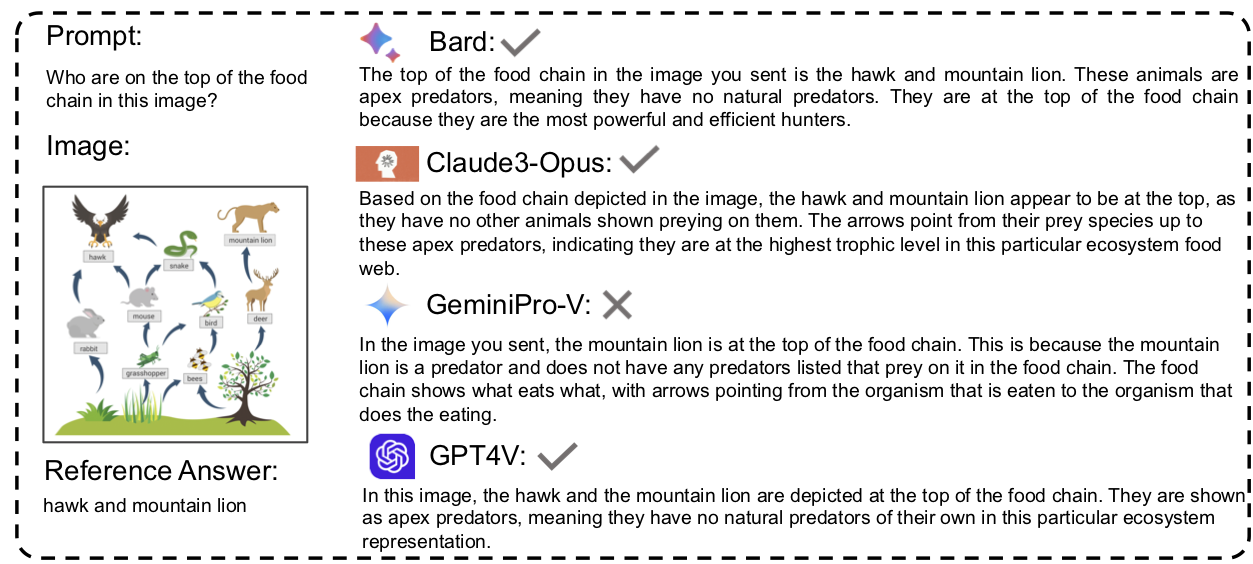}
		\caption{Education assistant case study I. Strong LVLMs understand the complex food chain well. Bard, Claude3-Opus, and GPT4V can answer the question correctly while GeminiPro-V only provides one of the answers.}
		\label{foodchain_test}
\end{figure*}

\begin{figure*}[h]
		\centering
		\includegraphics[width=1.0\linewidth]{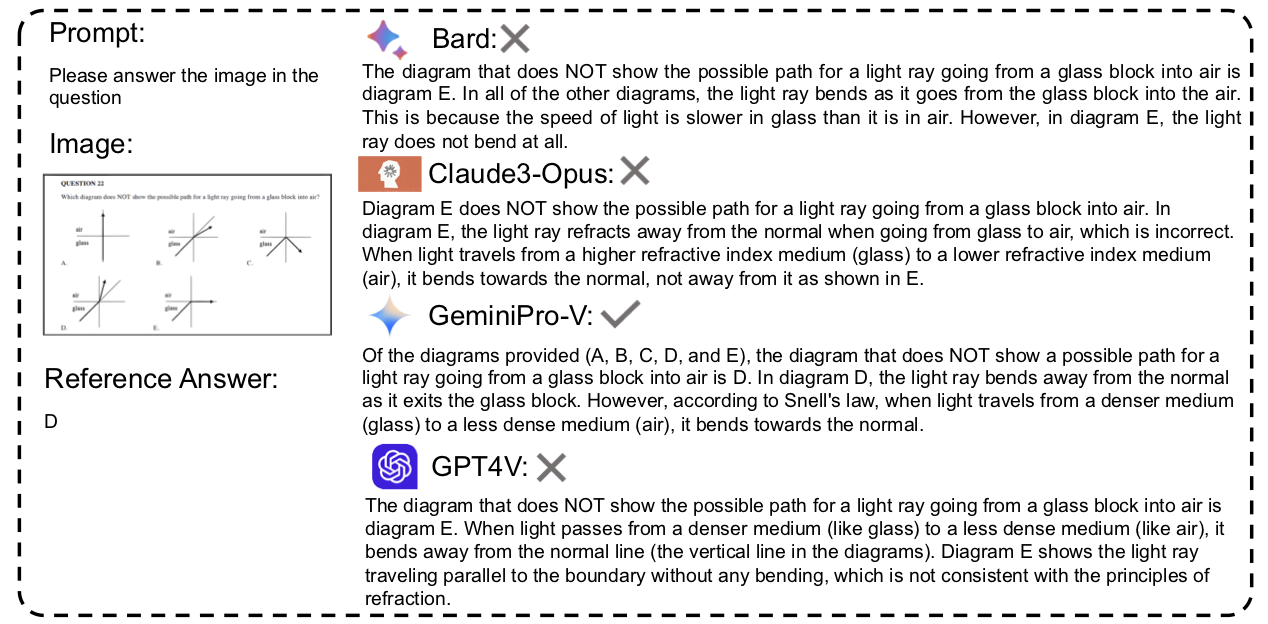}
		\caption{Education assistant case study II. Strong LVLMs struggle to effectively grasp and apply specialized physics knowledge.}
		\label{light_test}
\end{figure*}

\begin{figure*}[h]
		\centering
		\includegraphics[width=1.0\linewidth]{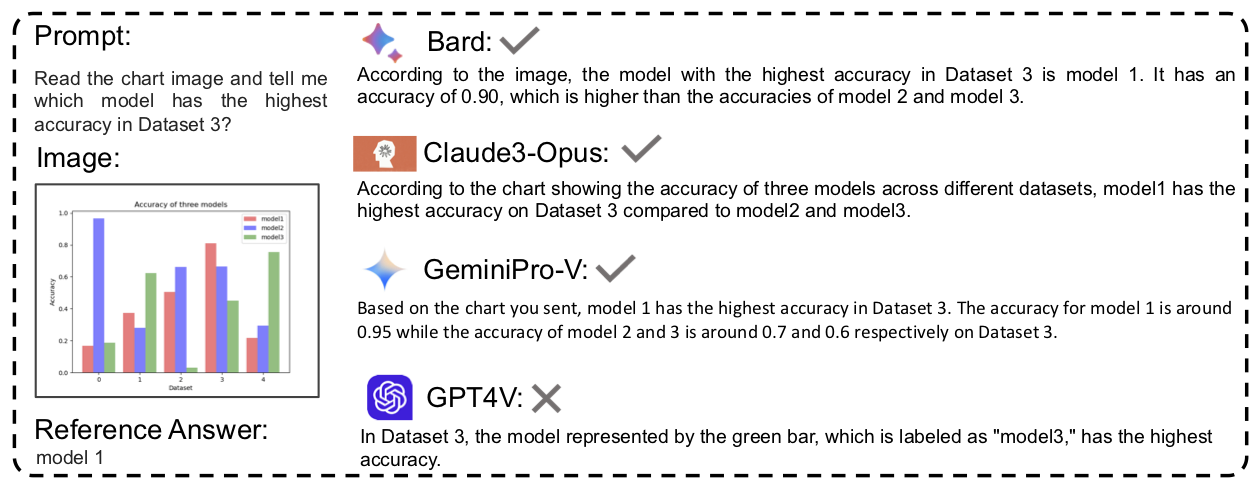}
		\caption{Document Analyst case study I. Strong LVLMs can understand elements and their relationships in the chart but struggle to extract numbers accurately.}
		\label{doc_chart_test}
\end{figure*}

\begin{figure*}[h]
		\centering
		\includegraphics[width=1.0\linewidth]{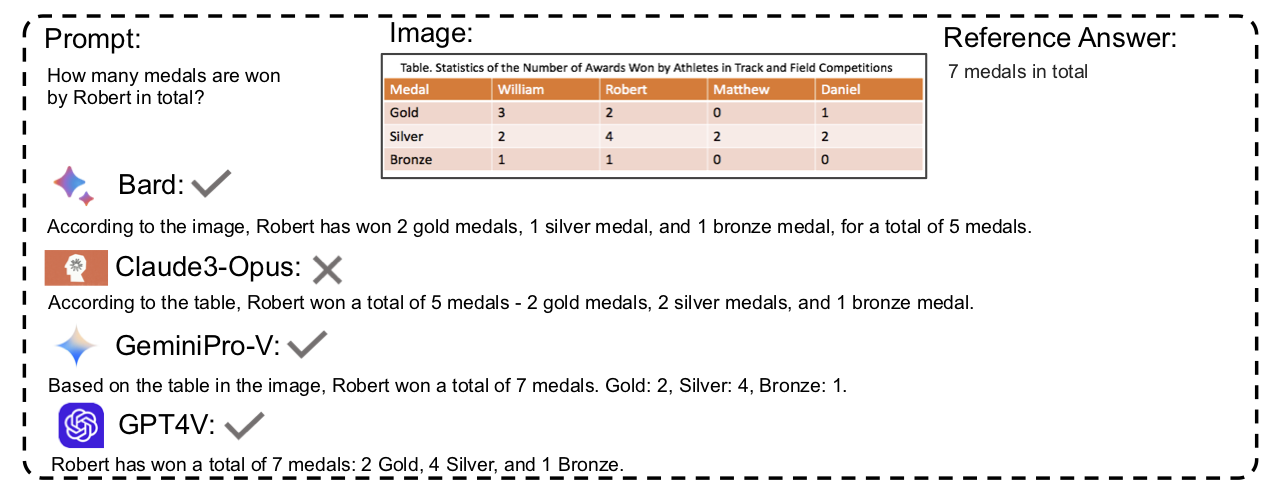}
		\caption{Document Analyst case study II. Strong LVLMs can understand and extract elements in the table image.}
		\label{doc_table_test}
\end{figure*}

\begin{figure*}[h]
		\centering
		\includegraphics[width=1.0\linewidth]{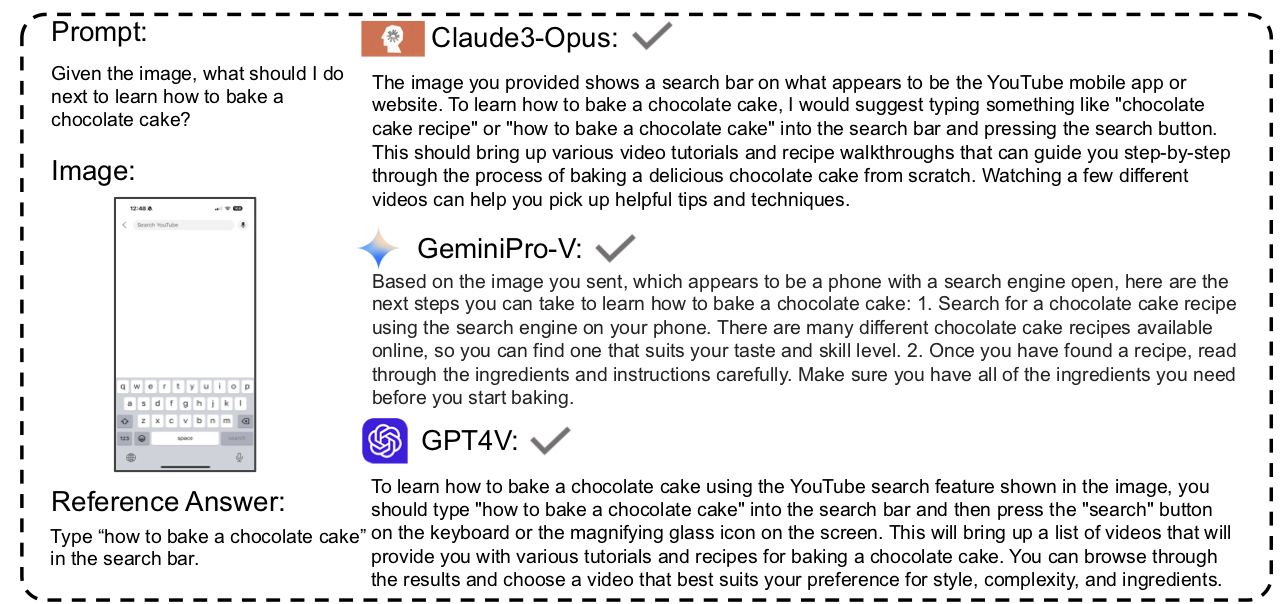}
		\caption{GUI Navigator case study I. Strong LVLMs can understand users' requests and the user interface of the phone.}
		\label{gui_type_test}
\end{figure*}

\begin{figure*}[h]
		\centering
		\includegraphics[width=1.0\linewidth]{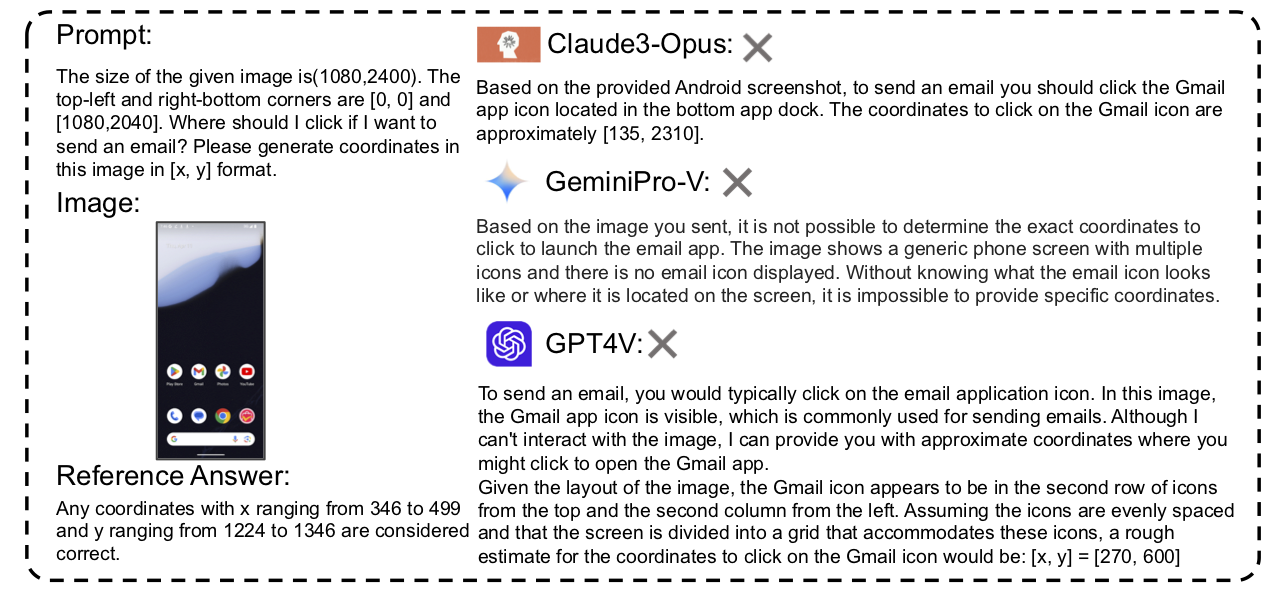}
		\caption{GUI Navigator case study II. Strong LVLMs can find the app icon for the user's request but cannot localize the app accurately.}
		\label{gui_click_test}
\end{figure*}

\begin{figure*}[h]
		\centering
		\includegraphics[width=1.0\linewidth]{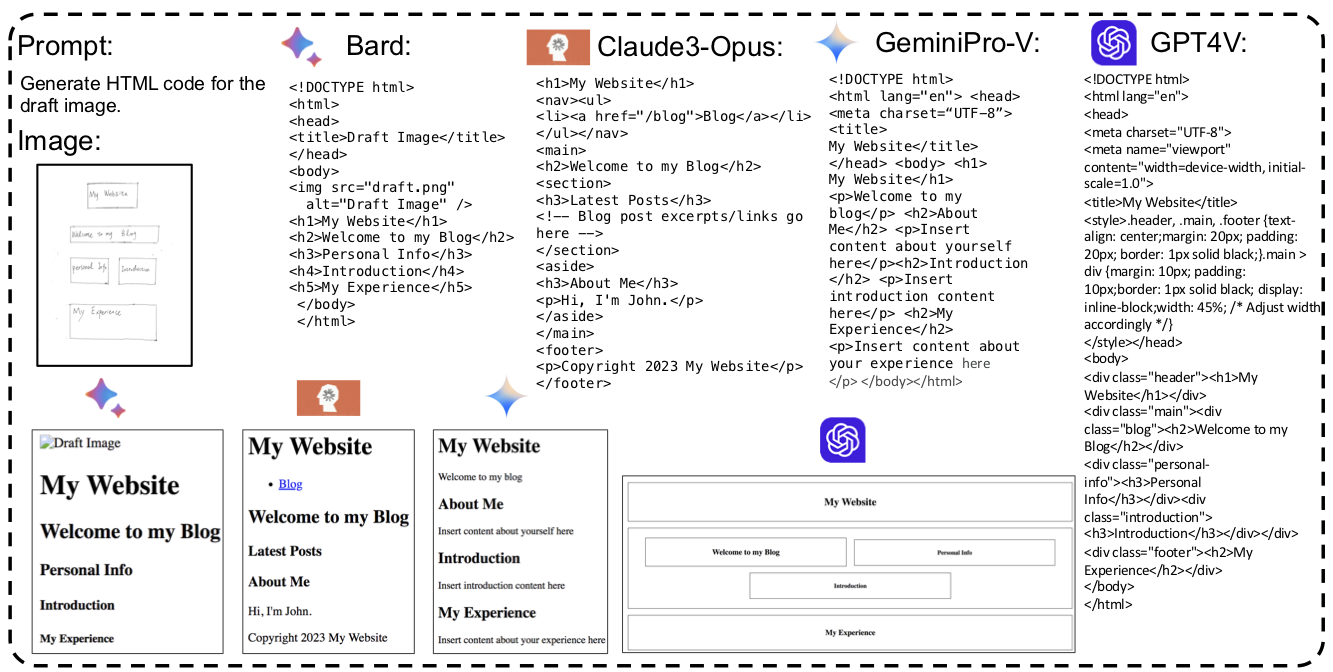}
		\caption{Code generator case study I. Strong LVLMs can generate code for user's request. GPT4V performs best because the generation of HTML code almost matches the layout of the user's draft.}
		\label{code_web_test}
\end{figure*}

\subsubsection{The sensitivity of CEE to randomness} We assess the sensitivity of ChatGPT Ensemble Evaluation (CEE) to random noise by visualizing the heatmap of score variances across 16 LVLMs. These results are based on evaluations conducted on 5 datasets (IC15, ImageNetVC shape, MSCOCO POPE adversarial, VCR MCI, and VSR) with three independent evaluations each. The results are reported in Fig. \ref{fig:llm-noise} where we can observe the small variance across multiple evaluation processes. Specifically, the average variance across all $16$  LVLMs of IC15, ImageNetVC shape, MSCOCO POPE adversarial, VCR MCI, and VSR are $1.39$, $1.20$, $1.29$, $1.04$, $1.15$, respectively. Our experiments indicate that CEE evaluation results are robust to random noise.

\subsubsection{The effect of different LLMs on CEE} We investigate the sensitivity of CEE to different LLM judges including ChatGPT (used by our CEE), GPT4, and language-only GeminiPro. To explore the influence of CEE with different LLM judges,  we calculate the correlation between overall scores of $16$ LVLMs using the proposed CEE evaluation. The results are reported in Fig. \ref{fig:llm-bias} where we see that the overall score sequences obtained by different LLMs are highly correlated with each other. It implies that the overall scores on TinyLVLM-eHub assessed by different LLMs are consistent with each other.

\subsection{Productivity Test of Bard, Claude3-Opus, GeminiPro-V, and GPT4V} \label{sec:exp-bard-demo}

In this section, we further assess the productivity of top-performing LVLMs, \textit{i.e.} Bard, Claude3-Opus, GeminiPro-V, and GPT4V). To this end, four types of multimodal applications are presented in Section~\ref{sec:exp-demo-edu} to Section~\ref{sec:exp-demo-code} including education assistant, document analyst, GUI navigator, and code generator.

\subsubsection{Education Assistant} \label{sec:exp-demo-edu}

We focus on LVLMs' ability as an education assistant to solve discipline VQAs. Firstly, we feed strong LVLMs with an image of the food chain and ask them to identify which animals are on the top of the food chain. As shown in Fig. \ref{foodchain_test}, Bard, Claude3-Opus, and GPT4V identify the hawk and mountain lion as the apex predators at the top of the food chain. They also provide correct reasoning steps that these animals are depicted without any natural predators in the image, and point out the arrows indicating their position at the top of the food web. While all three models agree on the answer, GeminiPro-V focuses solely on the mountain lion. It reasons that since the mountain lion is a predator and has no predators listed in the food chain, it must be at the top. This explanation considers only the mountain lion, not the hawk. 
Overall, strong LVLMs can understand the complex food chain well with minor flaws.

Secondly, LVLMs were evaluated using a physics problem concerning light propagation in a middle school context. As shown in Fig. \ref{light_test}, five diagrams were presented, and LVLMs were tasked with identifying the one that does not depict a plausible path for a light ray transitioning from glass to air. Only GeminiPro-V provided the correct answer. However, it erroneously stated that in diagram D, the light ray bends away from the normal upon exiting the glass block, indicating flawed reasoning. In reality, when light transitions from a denser medium (glass) to a less dense one (air), it bends away from the normal. The other models incorrectly identified answer E. Bard demonstrated a misunderstanding of light propagation laws, while Claude3-Opus presented an inaccurate law stating that light bends towards the normal when transitioning from a higher to a lower refractive index medium. Only GPT4V correctly comprehended the principle of refraction. Consequently, even strong LVLMs struggle to effectively grasp and apply specialized physics knowledge.

\subsubsection{Document Analyst}

Document understanding enhances the LVLM's ability to assist users, perform complex tasks, and contribute effectively in various domains such as education, research, customer service, and more.

We proceed with a visual question on the chart image. As shown in Fig. \ref{doc_chart_test}, the image depicts the accuracy of three models in five datasets. LVLMs are asked to find which model achieves the highest accuracy in Dataset 3. From responses, we see that Bard, Claude3-Opus and GeminiPro-V can identify that model 1 has the highest accuracy in Dataset 3. However,  Claude3-Opus does not provide reasoning details about the result and Bard and GeminiPro-V cannot extract the point in the chart accurately. In addition, GPT4V even gives the wrong output. These responses show that Strong LVLMs can understand elements and their relationships in the chart but struggle to extract numbers accurately.

We then provide these LVLMs with an easier case in document understanding. As shown in Fig. \ref{doc_table_test}, LVLMs are required to find the number of medals won by Robert in total with a table of medals won by different players provided. The table QA is easier than chart QA in Fig. \ref{doc_chart_test} because the correspondence between elements is shown in the structured table. As we see, Bard, GenminiPro-V, and GPT4V can answer the question correctly with the number of gold, silver, and bronze medals presented. Only Claude3-Opus recognizes the number of silver medals won by Robert inaccurately.

In conclusion, the evaluation highlights that while Strong LVLMs exhibit varying degrees of proficiency in document understanding tasks, challenges persist in accurately extracting and interpreting information, particularly from visual elements such as charts. While Bard, Claude3-Opus, and GeminiPro-V demonstrate the ability to identify relationships within the chart data, shortcomings in precise numerical extraction are evident across all models. Conversely, structured data in table format, as seen in the easier case, enables more accurate responses from Bard, GeminiPro-V, and GPT4V, indicating that structured presentation facilitates comprehension and extraction of relevant details. These findings underscore the importance of ongoing research and development efforts to enhance LVLMs' proficiency in document understanding, particularly in handling diverse data formats and extracting precise information for improved performance across various tasks and domains.

\subsubsection{GUI Navigator}
GUI navigation enables users to interact with LVLMs in an intuitive and user-friendly manner. Users can input queries, navigate through options, and visualize results more effectively through an LVLM agent. We assess whether strong LVLMs can be used as a good GUI navigator. Note that Bard is not included in the test because Bard is not accessible now.

Firstly, we feed LVLMs with a phone screenshot, and a search bar open on what appears to be the YouTube mobile app. The question is "Given the image, what should I do next to learn how to bake a chocolate cake?" As shown in Fig. \ref{gui_type_test}, Bard, Claude2-Opus, and GPT4V provide correct guidance. They accurately identify the search bar in the image and suggest using it to search for a chocolate cake recipe. While the level of detail provided differs. Claude3-Opus mentions following video tutorials, while GeminiPro-V emphasizes reading the recipe instructions. All the responses effectively address the user's query.

We proceed with a more difficult navigation task requiring fine-grained localization capability. As shown in Fig. \ref{gui_click_test}, Claude3-Opus and GPT4V analyze the screenshot correctly to answer the question about which app should be launched to send the email. But they give the wrong answer about where to click to launch the email app. GeminiPro-V even does not recognize the Gmail App in the image. Specifically, while Claude3-Opus identified the Gmail icon and its location in the bottom dock, it incorrectly provided coordinates to click on. The provided coordinates ([135, 2310]) do not point to the email app. GPT4V also faces the same issue.

In essence, while LVLMs show promise in understanding and responding to GUI elements in simpler scenarios, they encounter difficulties in tasks demanding precise localization and action identification within graphical interfaces. Continued research and refinement are necessary to enhance LVLMs' capabilities for seamless and accurate GUI navigation across diverse contexts and tasks.

\subsubsection{Code Generator}\label{sec:exp-demo-code}
Code generation enables translating high-level instructions or drafts into executable code, making them applicable across a wide range of domains such as software development, robotics, and automation.
We evaluate the capability of code generation of LVLMs with a draft of the webpage. 

As demonstrated in Fig. \ref{code_web_test}, all models can generate HTML code as required. The corresponding webpages are shown at the bottom of Fig. \ref{code_web_test}. We see that GPT4V performs best because the generation HTML code almost matches the layout of the user's draft. While the other three models can only generate simple HTML syntax and cannot achieve the webpage layout depicted in the sketch.

\section{Conclusion}

In this work, we propose a lightweight evaluation suite called TinyLVLM-eHub for Large Vision-Language Models (LVLMs). TinyLVLM-eHub comprehensively assesses various multimodal capabilities such as visual perception and embodied intelligence with quantitative evaluation. For a robust and accurate evaluation, we developed a new evaluation metric called CEE to assess the relationship between the reference answer and the answer predicted by LVLM. 
Through experiments, we demonstrate that CEE aligns better with human evaluation than the naive word match approach. By TinyLVLM-eHub, we reveal that closed-source API-Based LVLMs such as GPT4V and GeminiPro-V consistently outperform open-source LVLMs in various multimodal capabilities including visual perception, visual knowledge acquisition, visual reasoning, and embodied intelligence. Through various productivity tests, we also show that while top-performing LVLMs show promise in understanding elements in simpler scenarios, they encounter difficulties in tasks requiring expert domain knowledge, intricate image-text alignment, precise localization, and complex layout generation.

Although the evaluation in TinyLVLM-eHub is comprehensive, we only assess the boundary of multimodal capability for various LVLMs. Indeed, the evaluation of LVLM must also take into account other critical factors, such as content safety, political bias, and racial discrimination. These issues have become increasingly relevant due to the potential harm caused by biased or harmful content generated by these models. Therefore, it is crucial to thoroughly assess the ability of LVLM to produce safe and unbiased content that does not perpetuate harmful stereotypes or discriminatory attitudes. Furthermore, top-performing LVLMs have demonstrated remarkable proficiency in various multimodal capabilities, warranting a comprehensive investigation into specific aspects of their performance. Finally, TinyLVLM-eHub reveals the strengths and weaknesses of various LVLMs. Further exploration on developing LVLM should consider how to enhance the deep understanding of LVLMs in product-level scenarios.

\section*{Acknowledgments}
This paper is partially supported by the National Key R \& D Program of China No.2022ZD0160101 \& No.2022ZD0161000.

\section{References Section}
\bibliographystyle{unsrt}
\bibliography{egbib}



\section{Biography Section}
 

\begin{IEEEbiography}[{\includegraphics[width=1in,height=1.25in,clip,keepaspectratio]{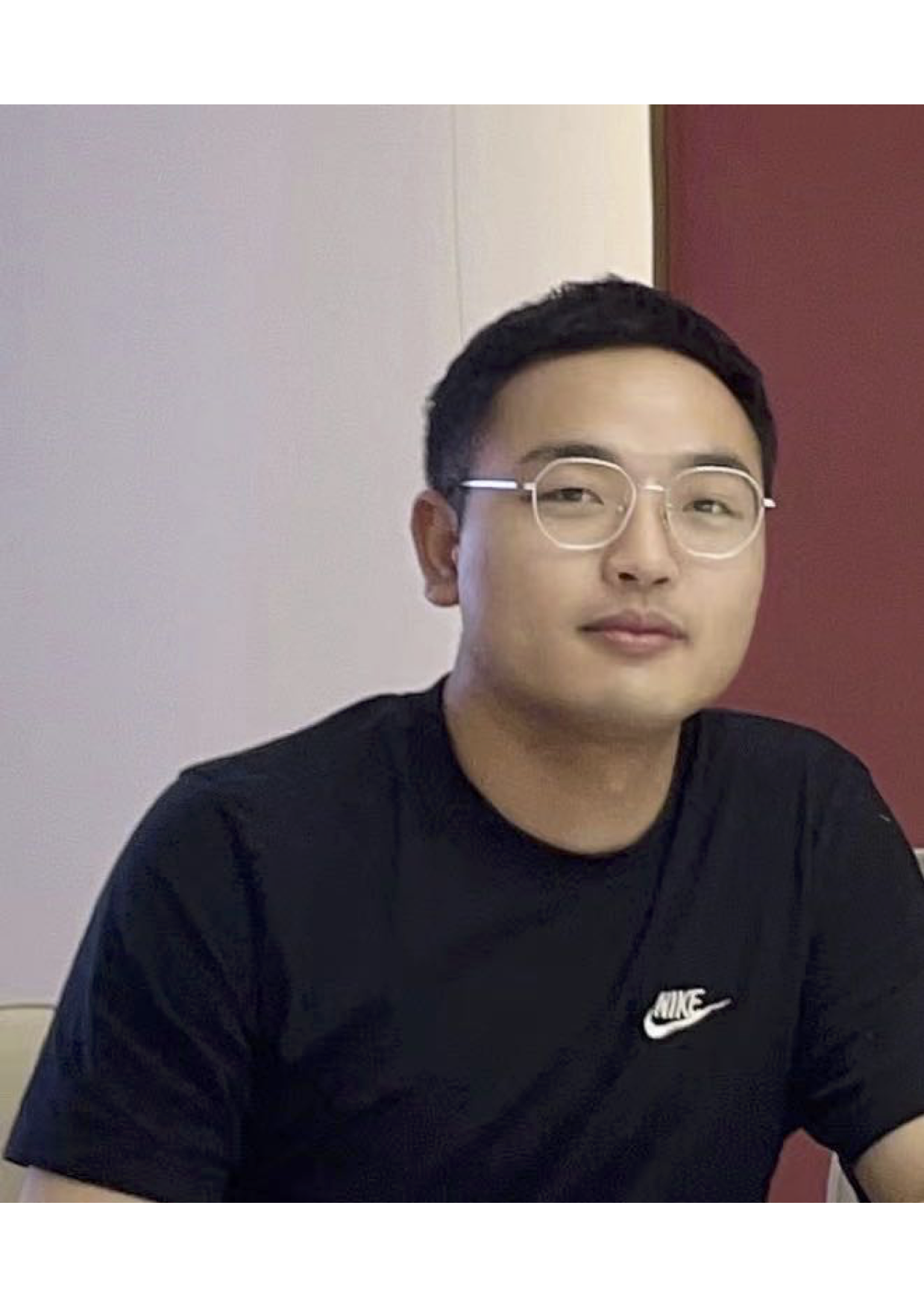}}]{Wenqi Shao}
 is a Young Scientist at the Shanghai Artificial Intelligence Laboratory. He completed his PhD in 2022 at the Multimedia Lab of the Chinese University of Hong Kong (CUHK), where he was supervised by Prof. Xiaogang Wang, Prof. Ping Luo, and Prof. Hongsheng Li. Prior to his doctoral studies, he obtained a bachelor's degree from the School of Mathematics at the University of Electronic Science and Technology of China (UESTC) in 2017.
His research interests revolve around multimodal foundation models, large language model compression, efficient transfer learning, and their applications in multimedia. Wenqi Shao is actively involved in the academic community and serves as a reviewer for several prestigious conferences, including CVPR, ICCV, ICML, NeurIPS, and ICLR.
\end{IEEEbiography}

\begin{IEEEbiography}[{\includegraphics[width=1in,height=1.25in,clip,keepaspectratio]{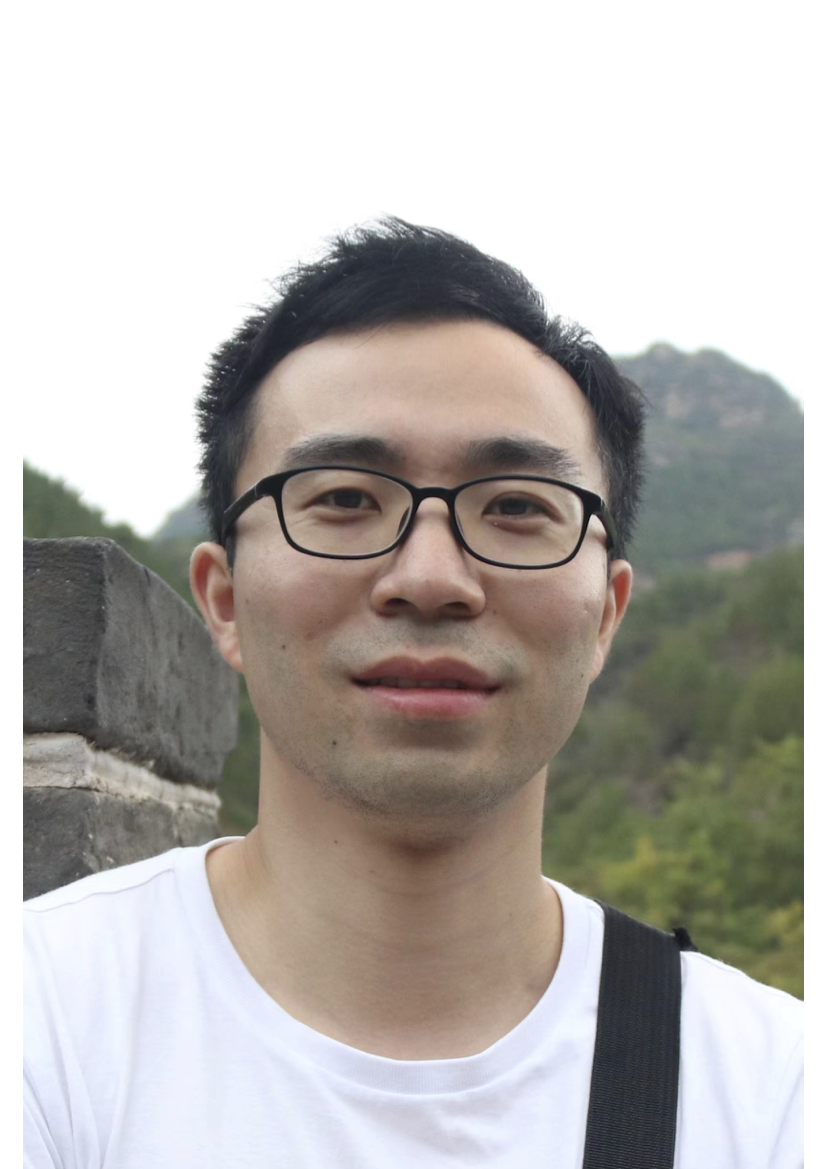}}]{Meng Lei}
 is a master's student at Peking University and received a BEng degree from The South China University of Technology. His research interest lies in computer vision and multimodal foundation models.
\end{IEEEbiography}

\begin{IEEEbiography}[{\includegraphics[width=1in,height=1.25in,clip,keepaspectratio]{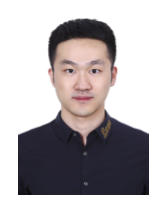}}]{Yutao Hu}
 received his B.S. degree in electronics and information engineering from Beihang University, Beijing, China in 2017, and he received his Ph.D. degree in the National Key Laboratory of CNS/ATM, School of Electronics and Information Engineering, Beihang University, Beijing, China. He is now working as a post-doctoral fellow in the University of Hong Kong. His research interests include machine learning and computer vision.
\end{IEEEbiography}

\begin{IEEEbiography}[{\includegraphics[width=1in,height=1.25in,clip,keepaspectratio]{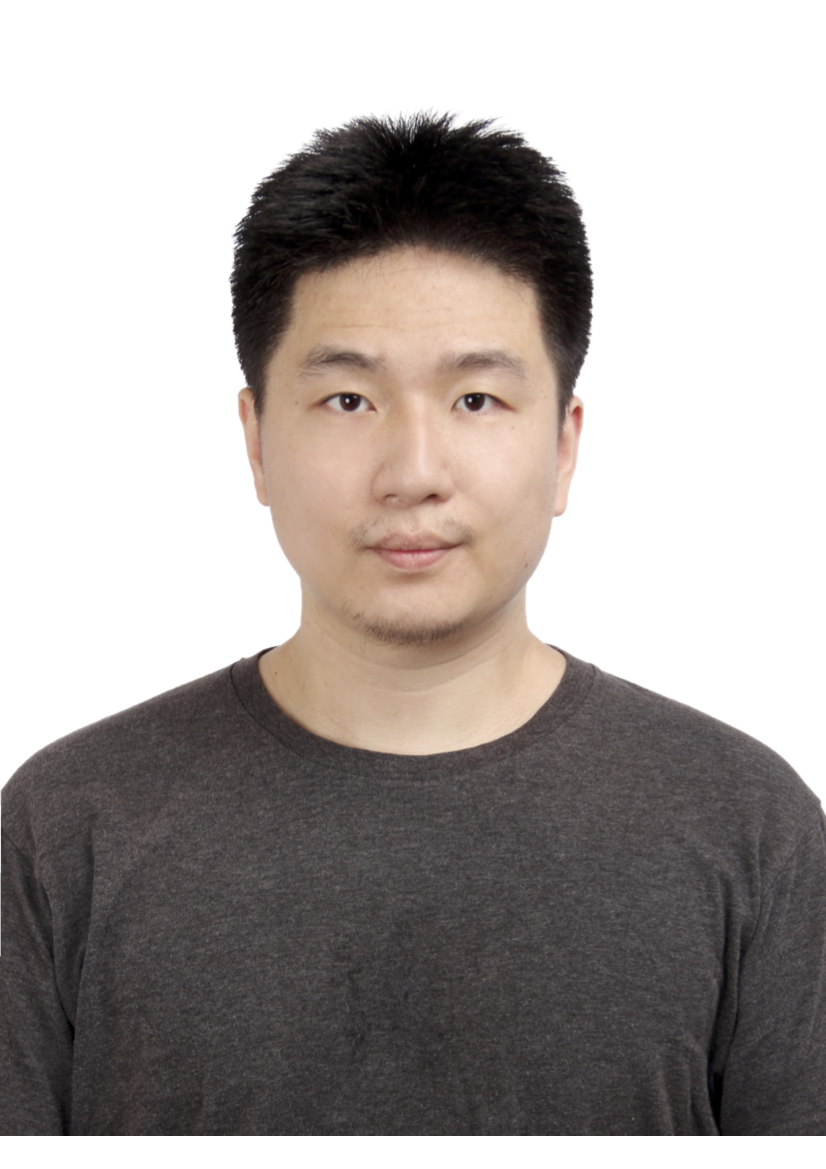}}]{Peng Gao}
 is a  Young Scientist at the Shanghai Artificial Intelligence Laboratory. He completed his Ph.D. in 2021 at the Multimedia Lab of the Chinese University of Hong Kong (CUHK), where he was supervised by Prof. Xiaogang Wang and Prof. Hongsheng Li. 
His research interests revolve around multimodal foundation models, efficient visual backbone design, self-supervised representation learning, and their applications in multimedia. Peng Gao is actively involved in the academic community and is a reviewer for several prestigious conferences, including CVPR, ICCV, ICML, NeurIPS, and ICLR.
\end{IEEEbiography}


\begin{IEEEbiography}[{\includegraphics[width=1in,height=1.25in,clip,keepaspectratio]{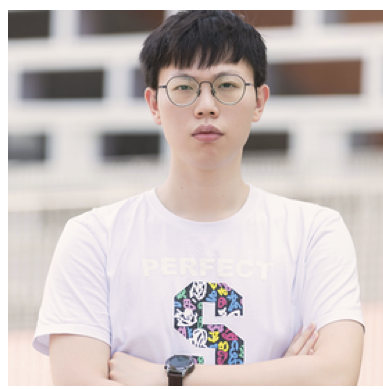}}]{Peng Xu}
 is currently a PhD student at the University of Hong Kong. Before this, he obtained his Bachelor's degree from the Southern University of Science and Technology, China. He is currently focusing on the inference techniques applied to large language models.
\end{IEEEbiography}


\begin{IEEEbiography}[{\includegraphics[width=1in,height=1.25in,clip,keepaspectratio]{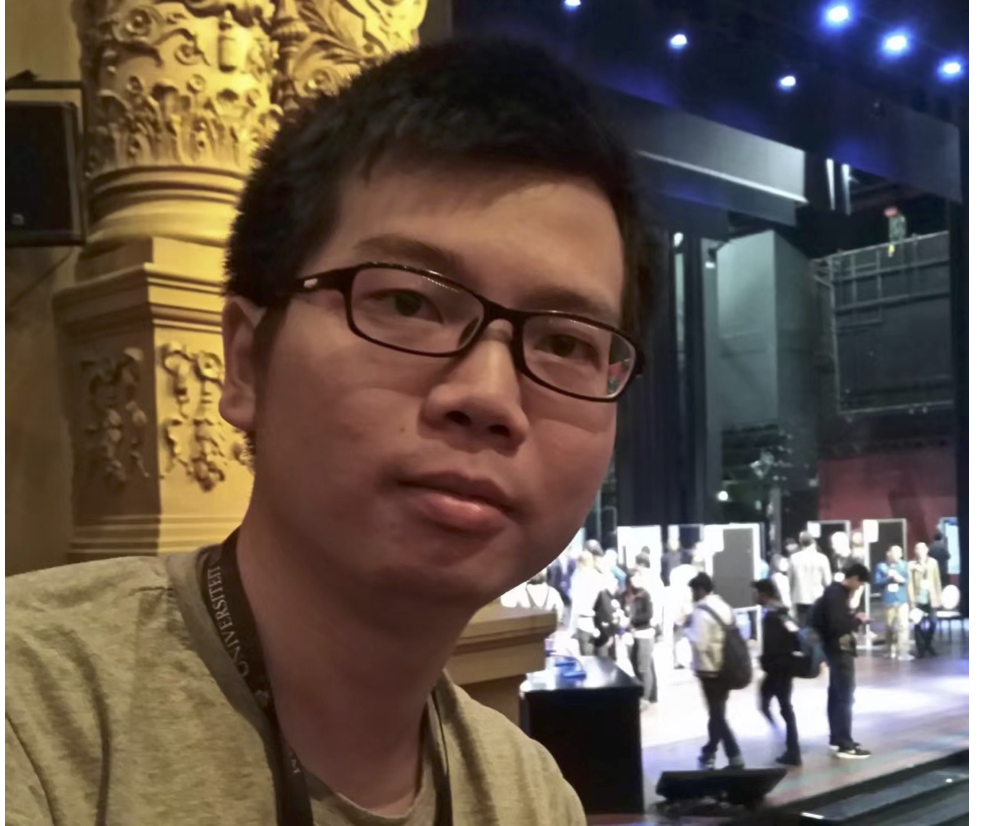}}]{ Kaipeng Zhang}
 received a B.S. degree in computer science from Donghua University, Shanghai, China, in 2016, an M.S. degree in computer science and information engineering from National Taiwan University, Taipei, Taiwan, in 2018, and a Ph.D. degree in Information Science from the University of Tokyo, Tokyo, Japan. He is currently a researcher at the Shanghai AI Laboratory. His research interests include face analysis, metric learning, and foundation models. He is a program committee member or reviewer for major international conferences and journals, such as CVPR, ICCV, ECCV, NeurIPS, ICML, and TPAMI. He is also a senior program committee member for IJCAI 2021.
\end{IEEEbiography}


\begin{IEEEbiography}[{\includegraphics[width=1in,height=1.25in,clip,keepaspectratio]{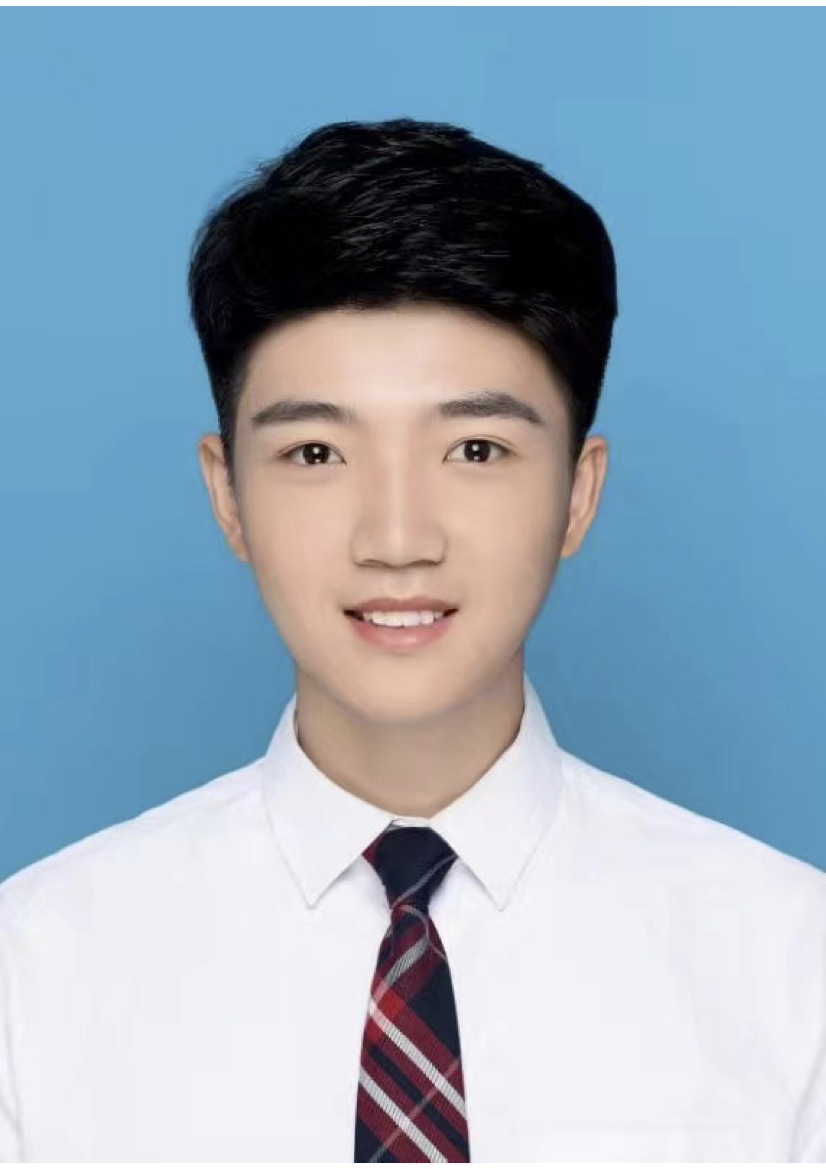}}]{Fanqing Meng}
received the B.S. degree in the School of Software Engineering, Tongji University, Shanghai, China. He is a first-year Ph.D. student in the School of Electronic Information and Electrical Engineering, at Shanghai Jiao Tong University, Shanghai, China. His current research interest focuses on the applications of computer vision as well as multimodal and transfer learning.
\end{IEEEbiography}



\begin{IEEEbiography}[{\includegraphics[width=1in,height=1.25in,clip,keepaspectratio]{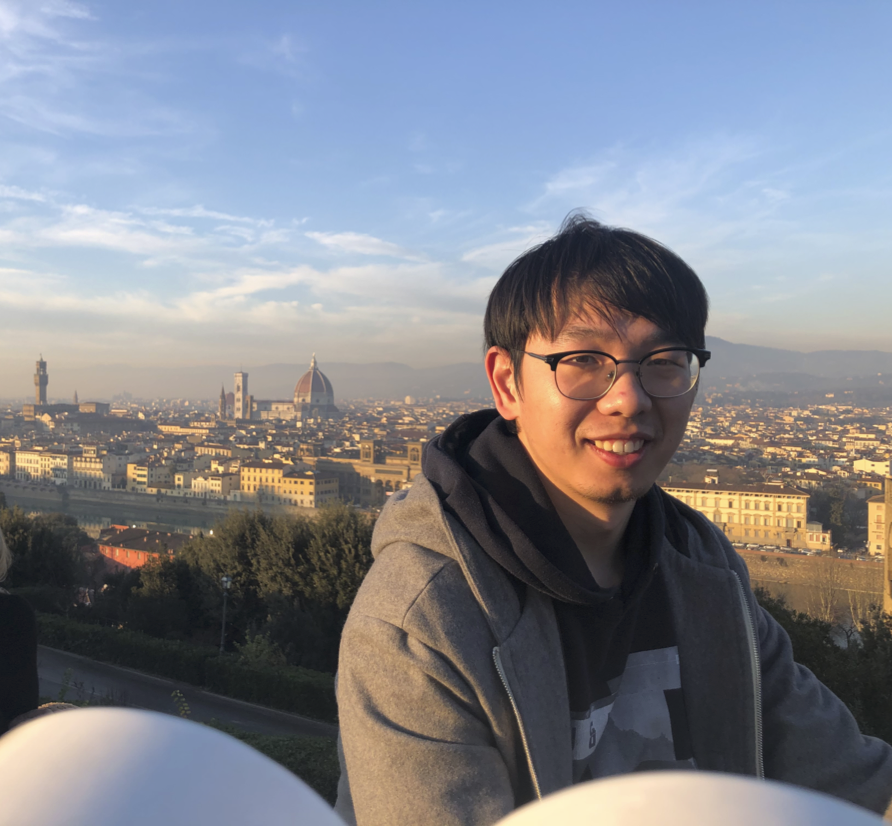}}]{Siyuan Huang}
 is a PhD student at Shanghai Jiao Tong University. He earned his M.Sc. degree from the Karlsruhe Institute of Technology and his B.Eng. degree from the Beijing Institute of Technology. His research interests lie in multimodal foundational models and robotics.
\end{IEEEbiography}


\begin{IEEEbiography}[{\includegraphics[width=1in,height=1.25in,clip,keepaspectratio]{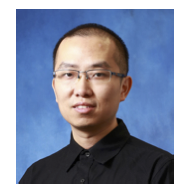}}]{Hongsheng Li}
received a bachelor's degree in automation from the East China University of Science and Technology, and master's and doctorate degrees in computer science from Lehigh University, Pennsylvania, in 2006, 2010, and 2012, respectively. He is currently an assistant professor in the Department of Electronic Engineering at The Chinese University of Hong Kong. His research interests include computer
vision, medical image analysis, and machine learning.
\end{IEEEbiography}


\begin{IEEEbiography}[{\includegraphics[width=1in,height=1.25in,clip,keepaspectratio]{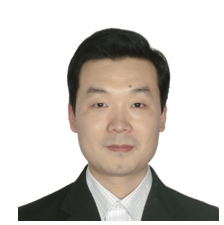}}]{Yu Qiao} (Senior Member, IEEE) is a professor
with Shanghai AI Laboratory and the Shenzhen
Institute of Advanced Technology (SIAT), Chinese Academy of Sciences. He has published
more than 600 articles in international journals
and conferences, including T-PAMI, IJCV, T-IP, TSP, CVPR, and ICCV. His research interests include computer vision, deep learning, and bioinformation. He received the First Prize of the
Guangdong Technological Invention Award, and
the Jiaxi Lv Young Researcher Award from the
Chinese Academy of Sciences. He is a recipient of the distinguished
paper award in AAAI 2021. His group achieved the first runner-up
at the ImageNet Large Scale Visual Recognition Challenge 2015 in
scene recognition, and the winner at the ActivityNet Large Scale Activity
Recognition Challenge 2016 in video classification. He served as the
program chair of IEEE ICIST 2014.
\end{IEEEbiography}


\begin{IEEEbiography}[{\includegraphics[width=1in,height=1.25in,clip,keepaspectratio]{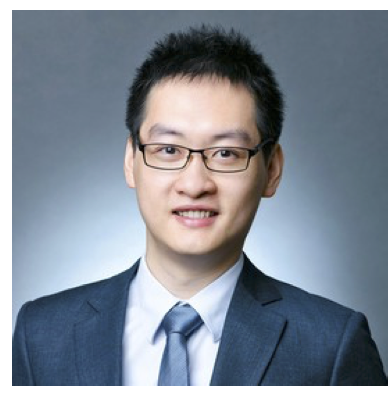}}]{Ping Luo}
 is an Associate Professor in the Department of Computer Science, The University of Hong Kong (HKU). He received his PhD degree in 2014 in Information Engineering, from the Chinese University of Hong Kong (CUHK), supervised by Prof. Xiaoou Tang and Prof. Xiaogang Wang. He was a Postdoctoral Fellow at CUHK from 2014 to 2016. He joined SenseTime Research as a Principal Research Scientist from 2017 to 2018. His research interests are machine learning and computer vision. He has published 100+ peer-reviewed articles in top-tier conferences and journals such as TPAMI, IJCV, ICML, ICLR, CVPR, and NIPS.
\end{IEEEbiography}


\vfill

\end{document}